\documentclass[pmlr,twocolumn,10pt]{jmlr} 

\usepackage{booktabs}
\usepackage{siunitx}

\usepackage[switch]{lineno}
\usepackage{multirow}
\usepackage{bm}
\usepackage{amsmath}

\theorembodyfont{\upshape}
\theoremheaderfont{\scshape}
\theorempostheader{:}
\theoremsep{\newline}

\jmlrvolume{333}
\jmlryear{2026}
\jmlrsubmitted{LEAVE UNSET}
\jmlrpublished{LEAVE UNSET}
\jmlrworkshop{Conference on Health, Inference, and Learning (CHIL) 2026} 

\title[Multi-dim EEG Eval]{A Multi-dimensional Framework for Evaluating Generalization in EEG Foundation Models}

\author{%
 \Name{Aditya Kommineni} \Email{akommine@usc.edu}\\
 \Name{Emily Zhou} \Email{emilyzho@usc.edu}\\
 \Name{Kleanthis Avramidis} \Email{avramidi@usc.edu}\\
 \Name{Tiantian Feng} \Email{tiantiaf@usc.edu}\\
 \Name{Shrikanth Narayanan} \Email{shri@usc.edu}\\
 \addr Signal Analysis and Interpretation Laboratory, University of Southern California, USA
}

\begin{document}

\maketitle

\begin{abstract}
Evaluating foundation models under appropriate adaptation settings is essential for understanding the quality and transferability of the learned representations.
Recent EEG foundation models have demonstrated promising transfer capabilities across tasks and datasets, motivating their growing use in neurotechnology and clinical applications.
However, these models are typically evaluated under full fine-tuning on well-curated downstream datasets, a setting that does not reflect biomedical domain constraints such as limited labeled data, reduced sensor coverage, or parameter-efficient adaptation.
In this work, we propose a multi-dimensional evaluation framework for assessing EEG models under realistic low-resource conditions.
Empirical analysis of both supervised EEG models and recent EEG foundation models, including LaBraM, CSBrain, and CBraMod, across 6 different datasets is performed under the proposed multi-dimensional evaluation framework.
We find that EEG foundation models consistently provide performance gains on long-context tasks such as sleep stage prediction and mental health state classification. 
In contrast, for short-window Brain Computer Interface style tasks, supervised models achieve comparable despite having substantially fewer parameters.
Additional analyses demonstrate that current foundation models provide limited robustness to short-window tasks and channel constrained settings.
Together, these findings motivate the use of multi-dimensional evaluation protocols that characterize model behavior under realistic use constraints.
\end{abstract}

\paragraph*{Data and Code Availability}
In this work, we use openly available datasets, Physionet MI~\citep{schalk2004bci2000, goldberger2000physiobank}, BCI Competition IV-2A~\citep{brunner2008bci}, Kaggle ERN~\citep{inria-bci-challenge}, TUEV~\citep{obeid2016temple}, Depression Classification (MDD MAL)~\citep{mumtaz2016mdd} and Sleep EDF datasets~\citep{kemp2000analysis}. All datasets are accessible on public platforms from respective dataset owners.
Associated code is available in Github repository ~\href{https://github.com/Uncertain-Quark/eegfm_eval_toolkit}{Link}.

\paragraph*{Institutional Review Board (IRB)}
Since all datasets are publicly available, this study did not require an IRB.

\section{Introduction}
\label{sec:intro}
Electroencephalography (EEG) plays a central role in clinical diagnostics and critical care. 
It is ubiquitously deployed in Epilepsy Monitoring Units~\citep{maganti2013eeg} for seizure monitoring and in polysomnography~\citep{rundo2019polysomnography} for diagnosing sleep disorders. 
Beyond clinical settings, recent developments in sensor hardware such as dry electrodes and portable, in-ear EEG~\citep{looney2012ear} yield promising directions for deployment of EEG wearable devices in ambulatory and real-world environments. 
Prior works have further demonstrated the utility of EEG in Brain-Computer Interfaces (BCIs)~\citep{lotte2018review} and neuro-rehabilitation~\citep{daly2008brain}. 
However, the idiosyncratic characteristics of EEG datasets, such as heterogeneous recording settings, low signal-to-noise ratio (SNR), and high annotation costs, continue to limit generalizability and performance in these applications~\citep{avramidis2025neural}.
\par 
In response to these challenges, there has been growing interest in extending the foundation model paradigm to biosignals~\citep{abbaspourazad2023large}, particularly EEG~\citep{zhou2025csbrain, wang2024cbramod, bendr,yang2023biot, kommineni2024knowledge, jiang2024large,avramidis2025neural}. 
Motivated by the rapid development of foundation models in audio, vision, and text, researchers hypothesize that large-scale pre-training could enable develop generalized representations robust to intrinsic variability and data scarcity inherent to EEG analysis. 
In these other domains, evaluation paradigms for foundation models have evolved beyond full fine-tuning to include low-resource adaptation~\citep{zhang2024low, dong2023efficient}, in-context learning~\citep{brown2020language}, and zero-shot transfer~\citep{elizalde2023clap}. 
This shift reflects the intended use of foundation models, namely leveraging large-scale pre-training to reduce downstream supervision and adaptation costs.
\par 
Despite this potential, the majority of existing evaluations of EEG foundation models rely on fine-tuning of \textit{all} model parameters using the entirety of downstream datasets~\citep{zhou2025csbrain, wang2024cbramod, yang2023biot, jiang2024large}. 
Evaluating these models solely under full fine-tuning further obscures their deployment utility whenever data, channels, or computational resources are limited.
\par 
To address this methodological gap, we propose a multi-dimensional approach to systematically evaluate EEG foundation models across dimensions directly corresponding to real-world constraints. In realistic clinical or ambulatory scenarios, models must often operate with limited subject data, reduced channel montages, short recording durations, or strict parameter budgets. Consequently, our framework evaluates generalization capabilities across model scale, data availability, and channel layout.
The contributions of this work are as follows:
\begin{itemize}
    \item We introduce a multi-dimensional evaluation framework, grounded in real-world deployment constraints, to assess the generalization behavior of EEG foundation models along three key axes: \textit{parameter}, \textit{sample}, and \textit{channel} efficiency.
    
    \item We implement parameter-efficient adaptation and low-resource training to probe the quality and utility of learned representations, quantifying the relative gains provided by large-scale pre-training over fully supervised baselines.
    
    \item We show that EEG foundation models exhibit improved sample efficiency on long-context temporal modeling tasks under severe data constraints, while achieving performance comparable to supervised models on short-window BCI tasks, highlighting task-dependent benefits and limitations of current pre-training strategies.
\end{itemize}
\section{Related Work}
\subsection{Foundation models for EEG}
Until recently, advances in representation learning for vision, audio, and text modeling were the primary motivators of foundation models for biosignals, and EEG in specific. 
Early work adapted self-supervised contrastive objectives~\citep{bendr}, demonstrating improved transferability across multiple disparate settings. 
The heterogeneous nature of neurophysiological recordings later shifted the research focus towards masked-reconstruction objectives~\citep{neurogpt, wang2024cbramod}, mainly considering transformer architecture~\citep{eegpt}. 
Still, scarcity of large-scale annotated data and high inter-subject variability of EEG acts as a performance bottleneck, such that more recent studies have sought to improve sample efficiency through methods, such as masked reconstruction~\citep{wang2024cbramod, zhou2025csbrain, wang2025eegmamba, ma2025codebrain}, knowledge-driven objectives~\citep{kommineni2024knowledge, jiang2025neurolm}, vector-quantized representation learning~\citep{jiang2024large, avramidis2025neural} and autoregressive modelling~\cite{liu2025echo}.
\subsection{Foundation Model Evaluation}
In adjacent domains, foundation models are primarily evaluated through transfer-oriented protocols, as full fine-tuning is often infeasible or costly. 
Common evaluation strategies include few-shot generalization, linear probing, and parameter-efficient adaptation, with comparisons against supervised models trained from scratch on each task. 
Among parameter-efficient methods, low-rank adaptation (LoRA) has emerged as a standard approach, updating only a small number of trainable parameters while keeping the model backbone frozen~\citep{hu2022lora}. 
Across vision, audio, and text, performance gains from foundation models are consistently most pronounced in low-resource regimes~\citep{brown2020language}, where fully supervised models tend to underfit or overfit.

In contrast, evaluation of EEG foundation models remains less standardized and is often limited to fine-tuning on a diverse set of downstream tasks. Most existing studies report improvements over supervised baselines via full end-to-end fine-tuning~\citep{jiang2024large,wang2024cbramod}, with comparatively fewer works restricting adaptation to task-specific heads or linear probes~\citep{lee2025large}. As a result, systematic evaluation protocols that isolate generalization and parameter efficiency across recording conditions remain limited~\citep{bomatter2025limited}.
\subsection{Heterogeneity in EEG Data}
A prominent challenge in developing and evaluating EEG foundation models lies in the heterogeneity of EEG data. 
EEG signals vary substantially across subjects, sessions, tasks, and recording setups, which reflects differences in cognitive state, electrode layout, and acquisition hardware.  
Prior studies have shown that inter-subject variability is a major limiting factor for EEG-based classifications. 
As a result, models trained on limited-scale multi-subject data frequently struggle to generalize to unseen individuals. 
Moreover, topological heterogeneity is another fundamental and pervasive challenge in EEG modeling, where each public dataset typically uses its own electrode layout. 
Some recent studies, like LUNA~\citep{doner2025luna} and MMM~\citep{yi2023learning}, have proposed a topological-invariant encoder to map arbitrary-sized channeled EEG data to a uniform and fixed-sized latent representation space. 
Finally, and importantly, task heterogeneity is also a frequently encountered barrier in developing EEG foundation models, as different cognitive states---such as sleep stages, motor imagery, or seizure events---are characterized by distinct spectral and temporal brain activities. 
As mentioned earlier, recent studies~\cite{zhou2025csbrain, wang2024cbramod, jiang2024large} aimed to improve cross-task generalization by scaling pre-training to larger volumes of diverse EEG data to learn robust and task-agnostic representations.
\section{Evaluating Generalization in EEG Foundation Models}
\begin{table*}[]
    \centering
    \scalebox{0.85}{
    \begin{tabular}{lcccccc}
    \toprule
     \textbf{Dataset} & \textbf{Task} &  \textbf{\# Classes} & \textbf{\# Subjects} & \textbf{Length (sec)} & \textbf{\# Channels} & \textbf{Sampling Rate (Hz)} \\
    \midrule
     \textbf{Physionet-MI} &Motor Imagery & 4 & 109 & 4 & 64 & 160\\
     \textbf{BCIC IV 2A} & Motor Imagery & 4 & 9 & 4 & 22 & 250\\
     \textbf{Kaggle ERN} & P300 & 2 & 26 & 1 & 56 & 200\\ 
     \textbf{MDD MAL} & Mental Health & 2 & 64 & 10 & 19 & 256\\ 
    
     \textbf{Sleep EDF} & Sleep Stage & 5 & 78 & 30 & 2 & 100\\
     \textbf{TUEV} & Events & 6 & 370 & 5 & 21 & 200\\
    \bottomrule
    \end{tabular}
    }
    \caption{Summary of dataset used for downstream evaluation of models. Sampling rate is reported in Hz and length in seconds. Supervised models were trained at the sampling rate of the recording whereas for SSL models, signals were resampled to 200Hz. Refer to Appendix~\ref{apd:datasets} for additional dataset details.}
    \vspace{-3mm}
    \label{tab:dataset_desc}
\end{table*}
\begin{table}
    \footnotesize
    \centering

    \scalebox{0.9}{
    \begin{tabular}{lccc}
        \toprule
        \multirow{2}{*}{\textbf{Model}} & 
        \multicolumn{1}{c}{\textbf{Param}} &
        \multicolumn{1}{c}{\textbf{Training}} &
        \multicolumn{1}{c}{\textbf{Training}} \\ 

        & 
        \multicolumn{1}{c}{\textbf{Size}} &
        \multicolumn{1}{c}{\textbf{Data Size}} &
        \multicolumn{1}{c}{\textbf{Objective}} \\ 

        \midrule
        
        LaBraM & 5.8M & 2.5+ kh & Masked Token Prediction \\ 
        CBraMod & 5M & 27.1 kh & Masked Reconstruction \\ 
        CSBrain & 9M & 9+ kh & Masked Reconstruction \\ 

        \bottomrule

    \end{tabular}
    }
    
    \caption{Overview of the parameter size, the pre-training data size, and the pre-training objective in EEG foundation models used for evaluation in this work.}
    \label{tab:eeg_foundation_model}
    \vspace{-3mm}
\end{table}
The core principle of foundation models is that large-scale unsupervised pre-training yields generalizable representations, enabling downstream adaptation under low-resource settings in a parameter efficient manner~\citep{brown2020language,radford2023robust, narayanswamy2024scaling, xu2025lsm, simeoni2025dinov3}.
The methodology for evaluating generalizability differs from one modality to other according to the idiosyncratic characteristics of the underlying signal.
For speech and text, models are evaluated based on their low-resource cross-domain and cross-lingual transfer capabilities~\citep{baevski2020wav2vec, wei2021finetuned}, whereas generalization in the domain of vision is characterized by the ability to extract semantically robust features across heterogeneous settings~\citep{radford2021learning, simeoni2025dinov3}. 
\par 
However, EEG presents a different set of challenges. 
Low signal to noise ratio, non-stationarity and large inter-subject and intra-subject variability often limit the ability of supervised models to generalize across settings. 
Moreover, large scale EEG data collections are constrained by participant fatigue, setup time and hardware limitations~\citep{sugden2023remote}.
As a result, generalization for EEG foundation models cannot be adequately captured by a single notion of transfer performance or dataset scaling.
We argue that generalization in EEG should be evaluated on a multidimensional axis, one that reflects realistic downstream constraints.
Through a combination of the three dimensions of parameter, sample, and channel efficiency, as well as corresponding details to operationalize them, we propose a generalized evaluation framework to assess the capabilities of EEG foundation models.
\paragraph{Parameter Efficiency}
While parameter efficiency is not unique to EEG models, the majority of existing EEG evaluations rely on full fine-tuning, where all parameters of a pre-trained model are adapted to the downstream task. As a result, full fine-tuning performance alone provides limited insight into the quality of the learned representations, since improvements may stem primarily from parameter updates rather than transferable features.

We study parameter efficiency under two complementary settings: \emph{linear probing} and \emph{parameter-efficient fine-tuning (PEFT)}. Linear probing evaluates the quality and generalizability of representations learned by the foundation model through training only a lightweight classifier on frozen features, while PEFT assesses the model’s ability to adapt to downstream tasks using a small number of additional or modified parameters.
To operationalize parameter efficiency, we define a relative performance metric ($PE_{S}$) that normalizes downstream performance under a given setting with respect to full fine-tuning:
\begin{equation}
\label{eq:1}
    PE_{S} = \frac{P_S - P_{chance}}{P_{FT}-P_{chance}}
\end{equation}
where $S \in \{\text{Linear Probe}, \text{PEFT}\}$, $P_{S}$ denotes the performance under setting $S$, $P_{\mathrm{FT}}$ corresponds to performance under full fine-tuning, and $P_{\mathrm{chance}}$ is the random-chance baseline. The performance metric $P$ is chosen according to the downstream task (e.g., balanced accuracy, macro-averaged F1).
The parameter efficiency score $PE_{S}$ measures the fraction of full fine-tuning performance achieved under a parameter-efficient regime, effectively disentangling representational quality from the benefits of extensive parameter adaptation. Models with high-quality and transferable representations are expected to exhibit values of both $PE_{\text{Linear Probe}}$ and $PE_{\text{PEFT}}$ close to 1.
\paragraph{Sample Efficiency}
Beyond parameter efficiency, a core criterion for evaluating foundation models is their ability to adapt under low-label regimes. Such settings are pervasive in EEG due to high annotation costs, limited data availability, and privacy constraints that restrict large-scale annotated data sharing.
To assess sample efficiency, we evaluate foundation models under varying total sample budgets, denoted by $S_{\text{total}}$, which corresponds to the total number of labeled training samples available for downstream adaptation. Experiments are conducted across representative values of $S_{\text{total}}$ under both linear probing and PEFT settings.

In addition to the total number of samples, an important consideration in low-resource EEG settings is the trade-off between participant diversity and the number of samples per participant. 
To capture this effect, we vary the number of subjects while keeping $S_{\text{total}}$ fixed. 
Hence, increasing the number of subjects leads to a proportionate decrease in number of samples per subject, enabling an explicit study of the balance between inter-subject diversity and per-subject data density under a fixed training budget.
While absolute performance under a given budget indicates whether a model can operate effectively in low-resource conditions, it does not reveal whether pre-training provides benefits beyond those achievable by supervised learning alone. 
To isolate the contribution of pre-training, we define sample efficiency as a relative performance measured with respect to a supervised baseline trained under the same total sample budget:
\begin{equation}
\label{eq:2}
SE^{D}_{S_{\text{Total}}} = \frac{P^{S_{\text{Total}}}_{D} - P_{\text{chance}}}{P^{S_{\text{Total}}}_{\text{Sup}} - P_{\text{chance}}}
\end{equation}
where $D$ denotes the adaptation setting (linear probe or PEFT), $P^{S_{\text{Total}}}_{D}$ is the downstream performance of the foundation model under budget $S_{\text{Total}}$, $P^{S_{\text{Total}}}_{\text{Sup}}$ is the performance of the supervised baseline trained with the same budget, and $P_{\text{chance}}$ denotes chance-level performance for the chosen evaluation metric.
Under this formulation, sample efficiency quantifies the relative advantage conferred by pre-training at a given data budget. Values of $SE^{D}_{S_{\text{Total}}} > 1$ indicate that the foundation model outperforms the supervised baseline under identical sampling constraints, while values near or below 1 suggest limited or no benefit from pre-training in that regime.
\paragraph{Channel Efficiency}
In most laboratory and clinical research settings, EEG is recorded using dense electrode montages, typically ranging from 64 to 256 channels. While such high-density recordings are essential for detailed spatial characterization and source localization of neural activity, they are impractical for many real-world and consumer-facing applications, where setup time, comfort, cost, and hardware constraints necessitate substantially fewer electrodes.
To evaluate whether EEG foundation models exhibit robustness under reduced sensing configurations, we study model performance under systematically constrained channel settings. Rather than randomly subsampling electrodes, we define two structured channel selection criteria that reflect realistic deployment scenarios.

First, we consider a \emph{sparse montage} setting, in which a fixed number of channels per cortical lobe are selected. This emulates low-density EEG caps while preserving coarse spatial coverage across the scalp. The number of channels per lobe is varied to progressively reduce the total channel count.
Second, we consider a \emph{lobe-restricted} setting, where electrodes are selected exclusively from a single cortical region (e.g., frontal, central, or midline). 
This setting reflects applications where electrodes are placed on localized brain regions due to task relevance or hardware limitations.
By evaluating performance under these controlled channel constraints, we assess the extent to which foundation models rely on dense spatial information versus their ability to leverage robust, transferable representations across brain regions under severe channel reduction. Unlike parameter and sample efficiency, channel efficiency is not summarized by a single scalar metric; instead, performance trends across channel configurations provide insight into spatial robustness and inductive biases learned during pre-training.
\section{Experimental Setup}
\label{sec:experiments}
\begin{table*}
    \centering
     \scalebox{0.85}{
    \begin{tabular}{lccccccc}
    \toprule
    \textbf{Model}  & \textbf{K-ERN} & \textbf{P-MI} & \textbf{IV-2A}& \textbf{TUEV} & \textbf{MDD MAL} & \textbf{Sleep EDF}   \\
    \midrule
    \texttt{Supervised}  \\
    \quad EEGNet (4K) & $62.74\pm2.09$ & $61.34\pm1.91$ & $54.96\pm7.05$ & $51.67\pm2.10$ & $86.89\pm3.88$  & $70.20\pm1.29$ \\
    \quad EEGNet Large (20K) & $64.94\pm2.23$ & $61.85\pm1.95$ & $58.14\pm7.07$ & $53.27\pm3.88$ & $84.98\pm6.82$ & $71.61\pm1.20$ \\
    \quad EEGNet Huge (110K) & $\underline{64.54\pm1.48}$ & $61.74\pm1.73$ & $\underline{58.76\pm6.60}$ & $52.58\pm1.82$ & $83.61\pm7.21$ & $71.14\pm1.36$ \\
    \quad EEGNex (50k)  & $\boldsymbol{65.00\pm1.38}$ & $\boldsymbol{65.58\pm 1.73}$ & $\boldsymbol{59.10\pm11.25}$ & $47.50\pm3.01$ & $86.21\pm8.36$ & $70.31\pm1.22$ \\
    \quad SparcNet (1M) & $61.70\pm1.30$ & $62.02\pm1.54$& $56.96\pm10.43$ & $52.10\pm3.51$ & $79.32\pm6.22$ & $71.01\pm2.64$ \\
    \midrule
    \texttt{Full Finetune} \\
    \quad LaBraM (5.8M) & $58.51\pm1.79$ & $57.25\pm1.55$ & $50.58\pm9.20$ & $\boldsymbol{57.58\pm1.91}$ & $\underline{88.72\pm3.12}$ & $\underline{72.86\pm1.22}$ \\
    \quad CBraMod (5M) & $61.47\pm1.25$ & $\underline{64.01\pm2.52}$ & $56.15\pm8.77$ & $\underline{54.70\pm2.34}$ & $83.08\pm6.69$ & $\boldsymbol{74.58\pm0.85}$ \\ 
    \quad CSBrain (9M)  & $63.19\pm1.60$ & $62.34\pm2.33$ & $55.27\pm8.39$  & $51.88\pm2.16$ & $\boldsymbol{88.81\pm5.41}$ & $71.17\pm1.17$ \\
    \bottomrule
    \end{tabular}
    }
    \caption{Downstream classification results for end-to-end training of supervised models and full-finetuning setting for foundation models. K-ERN, P-MI and IV-2A refer to Kaggle ERN, Physionet-MI and BCI Competition IV-2A dataset respectively.  Balanced Accuracy (BAC) metric is reported. For additional evaluation metrics, refer to Appendix~\ref{apd:full_results}. Parameter count for each model is reported within parentheses. Best and second-best performing metrics are in bold and underline, respectively.}
    \label{tab:all_datasets_full_training}
    \vspace{-3mm}
\end{table*}
\subsection{Datasets}
\label{sec:datasets}
To enable a holistic evaluation of performance across the proposed generalization dimensions, we conduct experiments on six representative EEG datasets spanning both short and long-duration tasks.
Short-window EEG tasks are often encountered in brain–computer interface (BCI) settings, where decisions are made from brief recording segments ($
\leq$5s). 
To reflect this regime, PhysioNet Motor Imagery (PhysioNet-MI), BCI Competition IV-2A (BCIC IV-2A), Kaggle Error-Related Negativity (Kaggle-ERN) and TUEV are included in this evaluation.  
PhysioNet-MI~\citep{schalk2004bci2000, goldberger2000physiobank} and BCIC IV-2A~\citep{brunner2008bci} are motor imagery datasets in which subjects perform real or imagined movements of specific body parts in response to visual cues.  
Kaggle-ERN~\citep{margaux2012objective, inria-bci-challenge} focuses on the detection of error-related responses, elicited when subjects attempt to spell a word using visually guided stimuli.
TUEV~\citep{obeid2016temple, harati2015improved} is a neurological event detection dataset.
\par
In contrast to short-window BCI tasks, an increasing number of clinically relevant EEG applications require modeling long temporal contexts. 
To capture this setting, we evaluate on MDD MAL~\citep{mumtaz2016mdd} and Sleep-EDF~\citep{kemp2000analysis} datasets which correspond to automated sleep staging and mental health assessment respectively.
\par 
Together, these datasets span diverse task structures, temporal scales, and clinical contexts, enabling a systematic assessment of generalization under varying resource and signal constraints.
To ensure robustness in results, BCIC IV-2A was evaluated in a Leave-One-Subject-Out (LOSO) setting and all other datasets were tested in a 5-fold, between-subjects cross-validation setup.
For additional dataset details please refer to Table~\ref{tab:dataset_desc} and Appendix~\ref{apd:datasets}.
\subsection{Data Preprocessing}
We adopted a uniform preprocessing pipeline for all supervised experiments which included notch filtering for powerline noise removal, band-pass filtering, re-referencing to the common average, and subsequent z-score normalization. 
For TUEV dataset, inter-quantile normalization was chosen over z-score to better preserve high-amplitude signal characteristics associated with artifact and epileptic events. 
For foundation models, we follow the preprocessing procedures reported in the respective original publications to ensure reproducibility of results.
\subsection{Model Architectures}
\label{sec:model_arch}
We evaluate 3 representative EEG foundation models as described in Table~\ref{tab:eeg_foundation_model}: LaBraM~\citep{jiang2024large}, CBraMod~\citep{wang2024cbramod}, and CSBrain~\citep{zhou2025csbrain}.  
For comparison against fully supervised approaches, the following EEG models are included: EEGNet~\citep{lawhern2018eegnet}, EEGNeX~\citep{chen2024toward}, and SparcNet~\citep{jing2023development}.
Refer to Appendix~\ref{apd:training} for further implementation details and model training.
To assess the parameter efficiency of foundation models during downstream adaptation, three fine-tuning strategies were considered:  
(1) \textit{Full fine-tuning}, in which all model parameters are updated;  
(2) \textit{Linear probing (LP)}, where the pre-trained backbone is frozen and only a linear head is trained; and  
(3) \textit{Parameter Efficient Fine-tuning (PEFT)}, which introduces a small number of trainable parameters to enable efficient adaptation.
These adaptation strategies allow us to disentangle performance gains arising from representation quality versus those attributable to increased parameter capacity.
\par 
For all linear probe evaluations, embeddings from the foundation models were flattened and a fully connected block with a single multi-head projection layer was used for fine-tuning. 
For parameter-efficient fine-tuning, low-rank adaptation (LoRA, by~\citet{hu2022lora}) modules were used for linear layers.
\begin{table*}
    \centering
    \begin{tabular}{lcccccccc}
    \toprule
    \textbf{Model}  & \textbf{K-ERN} & \textbf{P-MI} & \textbf{IV-2A} & \textbf{TUEV} & \textbf{MDD MAL} & \textbf{Sleep EDF}\\
    \midrule
    \texttt{Linear Probe} \\
    \quad LaBraM &  $0.74\pm.16$ & \textbf{$0.74\pm.01$} & \textbf{$0.82\pm.11$} & $0.64\pm.07$ & $0.95\pm.06$ & $0.84\pm.02$ \\
    \quad CBraMod& $0.68\pm.08$ & $0.69\pm.01$ & $0.62\pm.12$ & $0.55\pm.06$ & $0.96\pm.25$ & $0.69\pm.02$\\ 
    \quad CSBrain & $0.83\pm.05$ & $0.72\pm.03$ & $0.71\pm.10$ & \textbf{$0.87\pm.08$} & \textbf{$0.98\pm.11$} & \textbf{$0.85\pm.03$}\\
    \midrule
    \texttt{PEFT} \\
    \quad LaBraM & \textbf{$1.02\pm.25$} & $0.75\pm.01$ & $0.70\pm.07$ & $0.67\pm.07$ & $0.97\pm.06$ & $0.87\pm.05$\\
    \quad CBraMod & $0.93\pm.05$ & \textbf{$0.95\pm.02$} & $0.85\pm.18$ & $0.95\pm.08$ & $1.01\pm.15$ & $0.91\pm.03$\\ 
    \quad CSBrain & $0.89\pm.15$ &  $0.91\pm.04$ & \textbf{$1.05\pm.13$} & \textbf{$1.01\pm.08$} & \textbf{$1.01\pm.07$} & \textbf{$1.03\pm.02$}\\
    \bottomrule
    \end{tabular}
    \caption{Performance Efficiency ($PE_{S}$) values computed on Balanced Accuracy using Eq~\ref{eq:1} for LaBraM, CBraMod and CSBrain foundation models for the 6 evaluation datasets. For long-window tasks such as depression classification (MDD MAL) and sleep stage detection (Sleep EDF), foundation models provide higher parameter efficiency values compared to short-window tasks, indicating their representations being better aligned to long-context modeling. K-ERN, P-MI and IV-2A refer to Kaggle ERN, Physionet-MI and BCI Competition IV-2A dataset respectively.}
    \label{tab:pe_values}
\end{table*}
\subsection{Evaluation Metrics}
We evaluated all models using performance metrics that are well-established in EEG evaluation. We report balanced accuracy (BAC), F1-macro, and area under the receiver operating characteristic curve (AUROC)--for multiclass classification we instead report Cohen’s kappa. All the utilized metrics are described in detail in Appendix~\ref{apd:metrics}.
\section{Results}
\label{sec:results}
To establish an upper bound on downstream performance and contextualize subsequent resource-constrained analyses, we first evaluate supervised baseline models and fully fine-tuned foundation models (LaBraM, CBraMod and CSBrain) across the six evaluation datasets. 
During foundation model fine-tuning, weights were initialized from their respective pre-trained checkpoints.
Results in Table~\ref{tab:all_datasets_full_training} reveal complementary strengths across model classes. 
The benefits of foundation models are most pronounced in long-window clinical tasks. For example, in depression classification the top-performing foundation model achieves an absolute improvement of 4.2\% in Cohen’s Kappa, whereas on TUEV it achieves 4.0\% Macro-F1 improvement compared to the top-performing fully supervised model. 
However, foundation models are on par or slightly underperform supervised models in short-window BCI/ERN tasks, indicating limited effectiveness in modeling short temporal contexts.
Ablation experiments comparing the performance of supervised models at native sampling rate of corresponding dataset against the sampling rate of foundation models (200 Hz) indicate no noticeable difference, indicating that the performance discrepancy on short window tasks arises from model architecture and not sampling discrepancies (Appendix~\ref{apd:sampling_rate_ablations_appendix}).
Additional evaluation metrics for all experiments are reported in Appendix~\ref{apd:full_results}.
\subsection{Parameter Efficiency}
\begin{figure}
    \centering
    \includegraphics[width=0.5\linewidth]{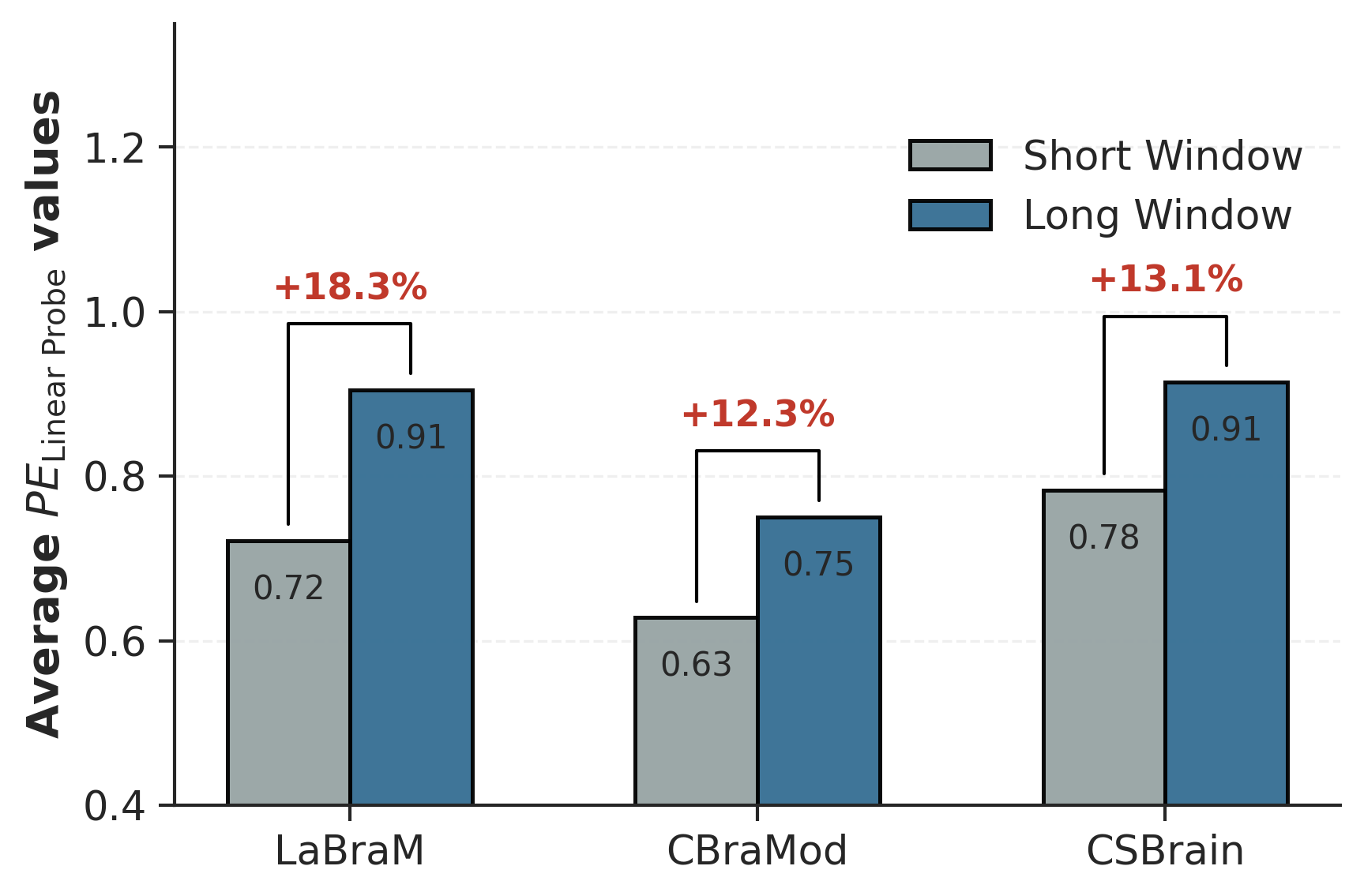}
    \caption{$PE_{\text{Linear probe}}$ values computed on BAC for LaBraM, CBraMod and CSBrain, aggregated based on task window length. Long-window tasks (Blue) parameter efficiency values are higher than short-window tasks (Grey) across all models.}
    \label{fig:linear_rel_probe}
    \vspace{-3mm}
\end{figure}

Evaluating foundation models under parameter constraints offers insights into the intrinsic quality of the pre-trained representations. 
As outlined in Section \ref{sec:model_arch}, we evaluate efficiency under two distinct regimes: a linear probe setting, which measures representation quality of frozen pre-trained features, and a PEFT setting, which tests the model's ability to adapt using only a fraction of the trainable parameters required for full fine-tuning. 
In Table~\ref{tab:pe_values} we report parameter efficiency values ($PE_{D}$) computed using balanced accuracy for linear probe and PEFT settings across evaluation datasets.
\paragraph{Linear Probe} CSBrain provides better representation quality over LaBraM and CbraMod models, as indicated by the higher linear probe parameter efficiency $PE_{\text{Linear probe}}$ values. 
This could be owing to better spatiotemporal information encapsulated in CSBrain, facilitated by inclusion of cortical lobe based embeddings~\citep{zhou2025csbrain}.
When comparing linear probe $PE$ across tasks, a distinct phenomenon is observed; across the three foundation models, short-window tasks (BCI IV-2A, Physionet-MI, Kaggle ERN, and TUEV) on average have lower values compared to long-window tasks (Sleep EDF, MDD MAL). 
Plot of average $PE_{\text{Linear probe}}$ between short-window and long-window tasks for the EEG foundation models in Figure~\ref{fig:linear_rel_probe} reveals the average $PE_\text{Linear probe}$ for short-window tasks to be at least 12\% lower (absolute) than for long-window tasks.
This performance gap suggests a fundamental difference in representation quality wherein foundation models provide better quality representations for longer window tasks.
\par 
\textbf{PEFT Performance}: 
For all parameter-efficient fine-tuning (PEFT) experiments, we employ LoRA with rank $r=4$ and scaling factor $\alpha=8$, resulting in only approximately 2–4\% of the full model parameters being trainable.
Despite this substantial reduction in trainable parameters, PEFT-adapted models achieve performance comparable to full fine-tuning on depression classification (MDD-MAL) and EEG event classification (TUEV), as shown in Table~\ref{tab:pe_values}.
In BCI tasks, PEFT substantially outperforms linear probing, with CBraMod and CSBrain attaining parameter efficiency values of up to 0.95. 
This indicates that updating the full parameter set (such as full fine-tuning setting) yields limited additional gains relative to the increase in model capacity, highlighting diminishing returns from full fine-tuning in these settings. 
These results suggest that PEFT provides a more effective trade-off between performance and trainable parameter count for EEG foundation models, particularly in resource-constrained adaptation scenarios.
\begin{table*}[h]
    \centering
    \begin{tabular}{llcccc}
    \toprule
     \textbf{Dataset} & \textbf{Model} & \multicolumn{4}{c}{\textbf{Total Sampling Budget ($S_{Total}$)}} \\
     \midrule
     &  & $\boldsymbol{50}$ & $\boldsymbol{100}$ & $\boldsymbol{150}$ & $\boldsymbol{200}$ \\
    \cmidrule(lr){3-6}
    \multirow{2}{*}{\textbf{BCIC IV-2A}} & Linear & $0.76 \pm 1.45$ & $0.40 \pm 0.44$ & $0.48 \pm 0.61$ & $0.69 \pm 0.93$ \\
     & LoRA & $0.56 \pm 0.89$ & $0.41 \pm 0.45$ & $0.16 \pm 0.81$ & $0.75 \pm 1.59$ \\
    \cmidrule(lr){1-2}
    \cmidrule(lr){3-6}
    && $\boldsymbol{240}$ & $\boldsymbol{480}$ & $\boldsymbol{960}$ & $\boldsymbol{1920}$ \\
    \cmidrule(lr){3-6}
    \multirow{2}{*}{\textbf{Sleep EDF}} & Linear & $2.91 \pm 1.60^{**}$ & $1.63 \pm 0.37^{**}$ & $1.37 \pm 0.29^{**}$ & $1.12 \pm 0.30^{*}$ \\
     & LoRA & $3.12 \pm 2.04^{**}$ & $1.79 \pm 0.51^{**}$ & $1.78 \pm 0.28^{**}$ & $1.31 \pm 0.36^{**}$ \\
    \bottomrule
    \end{tabular}
    \caption{Average Sampling Efficiency values at a given sampling budget across different test folds (using performance metric as Balanced accuracy in Eq.~\ref{eq:2}) for CSBrain (Foundation Model) Linear and LoRA settings at fixed total sampling budget compared to EEGNet Large (Supervised). Column headers denote budget for BCIC IV-2A and Sleep EDF respectively. Significance test for CSBrain performance being better than EEGNet Large are reported $^*p<0.05$, $^{**}p<0.001$.}
    \label{tab:se_values}
\end{table*}
\begin{figure*}
    \centering
    \includegraphics[width=0.9\linewidth]{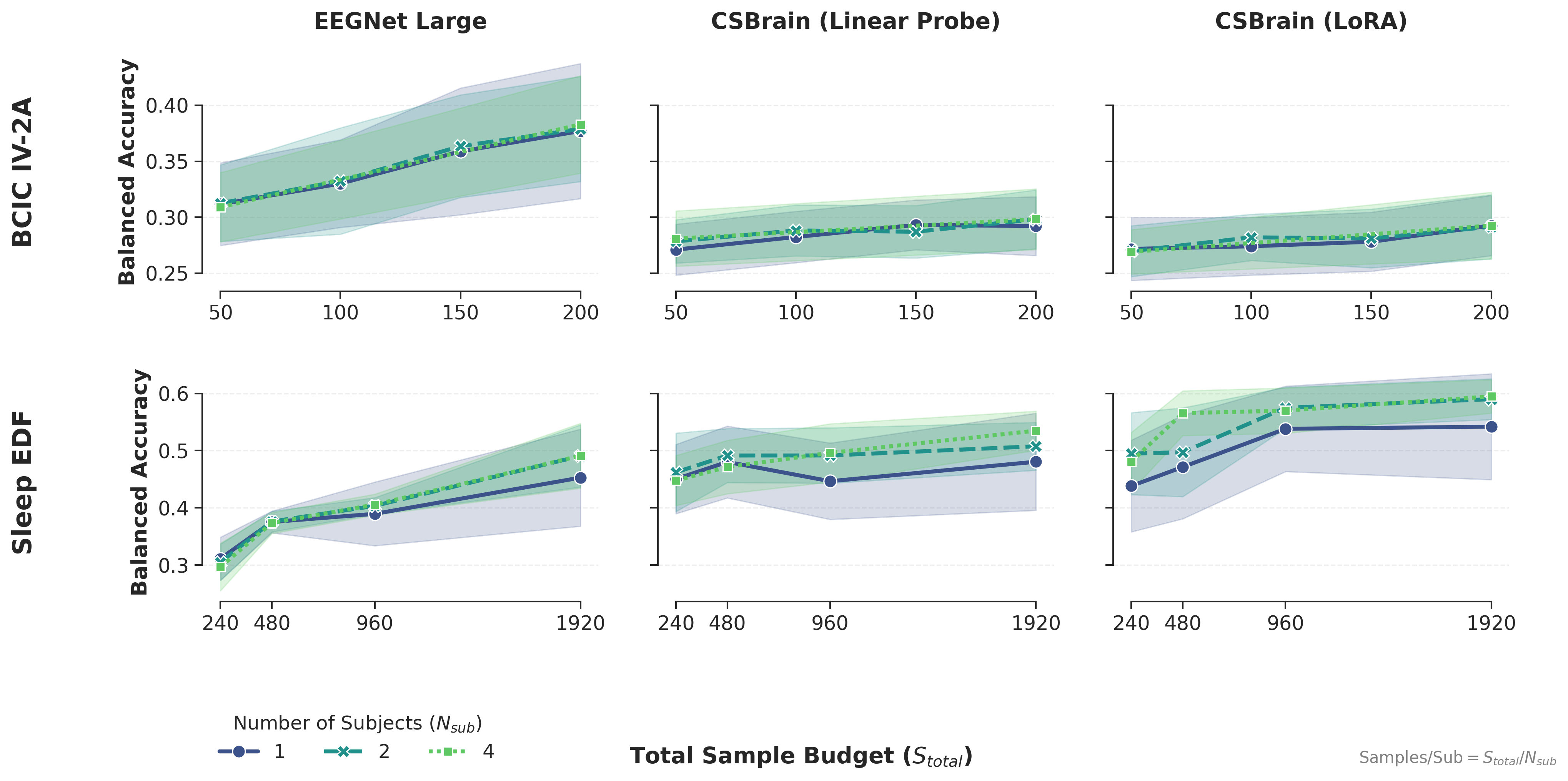}
    \caption{\textit{Sampling Efficiency}: Classification results for BCIC IV-2A and Sleep EDF datasets under fixed Total Sample Budget ($S_{total}$) for EEGNet Large and CSBrain under Linear Probe and LoRA. To ensure fixed $S_{total}$ under $N_{subjects}\in\{1,2,4\}$, number of samples per subject are adjusted accordingly. Each model run is repeated for 3 random seeds for 5 folds in Sleep EDF and Leave-one-subject-out setting for BCIC IV-2A.}
    \label{fig:sample_efficiency_csbrain}
\end{figure*}
\par 
\begin{figure*}
    \centering
    \includegraphics[width=0.8\linewidth]{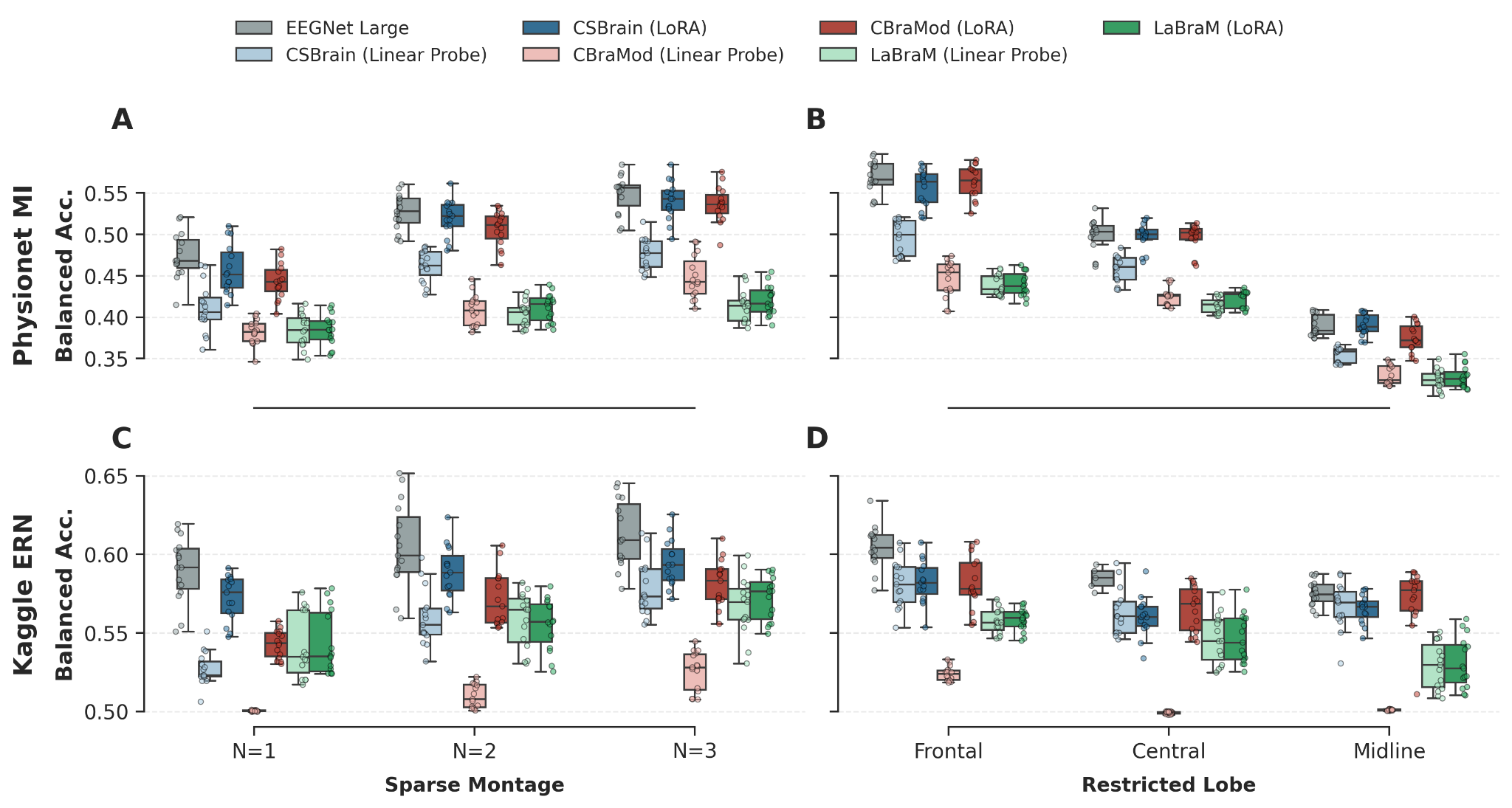}
    \caption{\textit{Channel Efficiency}: Boxplots of balanced accuracy of supervised (EEGNet Large) and foundation models (CSBrain, CBraMod, LaBraM) performance for Physionet MI and Kaggle ERN datasets under sparse montage and restricted spatial lobe conditions. (A,C) correspond to classification performance under uniformly reducing channels per corticol lobe (central, frontal, temporal, parietal and occipital) to $N\in\{1,2,3\}$. (B,D) correspond to classification performance of models under selecting frontal, central and midline lobes.}
    \label{fig:channel_ablations}
\end{figure*}
\subsection{Sampling Efficiency}
To study model behavior under low-resource settings, we evaluate performance under a fixed total sample budget ($S_{\text{Total}}$), representing the size of the training set used for downstream classification. Experiments are conducted on two representative datasets: a short-window motor imagery task (BCIC IV-2A) and a long-window sleep staging task (Sleep EDF).
Sampling efficiency values ($SE^D_{S_{\text{Total}}}$) are computed using EEGNet-Large as the supervised baseline, while CSBrain is evaluated under linear probing and LoRA fine-tuning ($r=1$) settings. For BCIC IV-2A, total sample budgets of 50 ($\sim$5 minutes of data), 100, 150, and 200 samples are considered. For Sleep EDF, budgets are set to 240 ($\sim$2 hours), 480, 960, and 1920 samples. In all cases, samples are class-stratified within each subject to control for class imbalance.
\par 
Sampling efficiency results in Table~\ref{tab:se_values} reveal distinct behaviors across the two tasks. For Sleep EDF, CSBrain under both linear probing and LoRA adaptation consistently outperforms EEGNet-Large. Moreover, the relative performance gains increase as the total sample budget decreases, with average $SE^D_{S_{\text{Total}}}$ values of 2.91 for linear probing and 3.12 for LoRA at 240 samples ($\sim$2 hours of data). As more training data becomes available, the magnitude of $SE^D_{S_{\text{Total}}}$ decreases, yet remains statistically significantly greater than one, indicating sustained advantages for foundation model representations.
In contrast, for the BCIC IV-2A dataset, $SE^D_{S_{\text{Total}}}$ values remain below one across all sampling budgets, highlighting the limited effectiveness of current EEG foundation models under low-sample regimes for short-window BCI tasks.
\par 
Low-resource settings in the context of EEG arise along two distinct axes, limited number of subjects and a limited number of samples per subject. 
As shown in Fig.~\ref{fig:sample_efficiency_csbrain}, under severe data constraints, increasing subject diversity does not compensate for a limited total sample budget. 
Instead, performance is primarily governed by the total number of training samples ($S_{\text{total}}$), largely independent of how those samples are distributed across subjects.
While prior works~\citep{bomatter2025limited} indicate some improvement in downstream performance with increase in subject diversity, their evaluations were conducted under much larger training sample set, which could indicate different behavior depending on scale of training data.
Additional Sample Efficiency results for MDD-MAL dataset are provided in Appendix~\ref{apd:mdd_mal_sample_efficiency}.
\subsection{Channel Efficiency}
To study the robustness of EEG foundation models under reduced channel availability, we consider two complementary channel-reduction strategies, as described in Section~\ref{sec:experiments}: a sparse-montage setting and a lobe-restricted setting. In the sparse-montage setting, the number of channels per cortical lobe (central, frontal, temporal, parietal, and occipital) is uniformly reduced to $N \in \{1,2,3\}$, resulting in total channel counts of $\{5, 10, 15\}$, respectively. This setup emulates scenarios where sparse yet spatially distributed electrode coverage is required. In the lobe-restricted setting, channels are confined to specific regions of the montage—namely frontal, central, and midline—reflecting practical constraints where only partial scalp coverage is available.
Both channel-reduction settings are evaluated on two BCI tasks: motor imagery classification using the Physionet MI dataset and error-related negativity detection using the Kaggle ERN dataset. We compare the supervised baseline EEGNet-Large with the foundation models CBraMod, CSBrain and LaBraM under parameter-efficient adaptation via linear probing and LoRA.
\par 
Results in Fig.~\ref{fig:channel_ablations} show that foundation models perform on par with, or slightly worse than, EEGNet-Large across all channel configurations. This suggests that current EEG foundation models do not yet exhibit enhanced robustness to reduced or regionally restricted channel inputs, indicating the need for improved pre-training strategies that more explicitly capture channel-specific and spatial inductive biases.
Comparing the two channel selection strategies, uniform per-lobe sampling yields more stable performance across tasks despite introducing sparsity. 
In contrast, restricting channels to a single region can lead to pronounced performance degradation when task-relevant information is distributed across lobes. 
This effect is particularly evident for Physionet MI when using only midline electrodes, as motor imagery signals are known to be strongly lateralized as seen in Fig.~\ref{fig:channel_ablations}B.
\section{Discussion}
\textbf{Tokenization Granularity could limit short window generalizability in Foundation Models.}
Despite strong performance on long-sequence inputs ($>5$s), foundation models underperform supervised baselines on short-window tasks such as Physionet-MI, BCIC IV-2A, and Kaggle-ERN, even with substantially higher parameter counts.
This discrepancy could be attributed to the tokenization strategies employed by foundation models. Specifically, EEG signals are typically segmented into coarse one second time window patches and processed using transformer-based backbones in order to enable modeling of long temporal contexts during pre-training~\citep{zhou2025csbrain, wang2024cbramod}.
While such tokenization is necessary to model long-context temporal dependencies, as reflected in the observed performance gains in sleep stage prediction and depression classification, it might limit the model's ability to capture fine-grained, transient neural dynamics, such as ERP-like responses, that are critical for short-window classification tasks.
\paragraph{EEGNeX Performance in Short-Window BCI Tasks.}
EEGNeX consistently outperforms other models on short-window BCI tasks. Prior work attributes this advantage to the use of strided convolutions~\citep{chen2024toward}, which may enable more effective aggregation of local temporal context within brief signal segments compared to standard CNN designs and transformer-based foundation models.
\paragraph{Limitations} This work primarily focuses on masked reconstruction based EEG foundation models (LaBram, CBraMod and CSBrain) trained under varying pre-training data sizes, parameter counts and compute budgets. 
Future work should consider evaluating EEG foundation models trained under identical settings, that enables isolating impact of individual pre-training axes on downstream performance.
Additionally, while we analyze sample efficiency under constrained data budgets on representative datasets, we do not provide a comprehensive evaluation across a broader range of tasks and recording conditions.
In particular, the interaction between inter-subject and intra-subject variability and different data sampling strategies in low-resource EEG settings remains an open problem that warrants further study.
\section{Conclusion}
In this work, we propose a multi-dimensional framework to evaluate EEG foundation models.
By systematically assessing performance across parameter, sample, and channel constraints, we uncovered distinct trade-offs across these dimensions in current EEG foundation models.
Our empirical analysis across six diverse datasets reveals that the utility of current EEG foundation models is highly task-dependent. We observed that these models excel in long-context tasks, such as sleep staging and mental health assessment, where they demonstrate performance gains over supervised baselines under full-finetuning and resource constrained settings.
However, this advantage does not yet extend to short-window BCI tasks or scenarios with channel constraints.
In these settings, supervised models (like EEGNet, EEGNeX) remain competitive or superior, indicating that current tokenization and pre-training objectives may not adequately capture the fine-grained spatiotemporal features necessary for tasks like motor imagery.
Future research could focus on developing pre-training objectives that explicitly encode short-window dynamics and channel invariance to bridge the gap between high-resource research benchmarks and realistic, low-resource deployment.
\section*{Generative AI Use Disclosure}
Large Language Models (LLMs) were used for refining text during preparation of manuscript.
Additionally, LLMs were employed to generate parts of the code implementation.
For LLM-generated content in the manuscript and code, the content was verified by the authors prior to inclusion.

\acks{
This study was sponsored by the Defense Advanced Research Projects Agency (DARPA) under cooperative agreement No. N660012324006. The content of the information does not necessarily reflect the position or the policy of the Government, and no official endorsement should be inferred.
}

\bibliography{main}

@article{bomatter2025limited,
  title={Is Limited Participant Diversity Impeding {EEG-based} Machine Learning?},
  author={Bomatter, Philipp and Gouk, Henry},
  journal={arXiv preprint arXiv:2503.13497},
  year={2025}
}

@article{yi2023learning,
  title={Learning topology-agnostic {EEG} representations with geometry-aware modeling},
  author={Yi, Ke and Wang, Yansen and Ren, Kan and Li, Dongsheng},
  journal={Advances in Neural Information Processing Systems},
  volume={36},
  pages={53875--53891},
  year={2023}
}

@article{doner2025luna,
  title={LUNA: Efficient and topology-agnostic foundation model for {EEG} signal analysis},
  author={D{\"o}ner, Berkay and Ingolfsson, Thorir Mar and Benini, Luca and Li, Yawei},
  journal={arXiv preprint arXiv:2510.22257},
  year={2025}
}

@article{lawhern2018eegnet,
  title={EEGNet: a compact convolutional neural network for EEG-based brain--computer interfaces},
  author={Lawhern, Vernon J and Solon, Amelia J and Waytowich, Nicholas R and Gordon, Stephen M and Hung, Chou P and Lance, Brent J},
  journal={Journal of neural engineering},
  volume={15},
  number={5},
  pages={056013},
  year={2018},
  publisher={iOP Publishing}
}

@article{jing2023development,
  title={Development of expert-level classification of seizures and rhythmic and periodic patterns during {EEG} interpretation},
  author={Jing, Jin and Ge, Wendong and Hong, Shenda and Fernandes, Marta Bento and Lin, Zhen and Yang, Chaoqi and An, Sungtae and Struck, Aaron F and Herlopian, Aline and Karakis, Ioannis and others},
  journal={Neurology},
  volume={100},
  number={17},
  pages={e1750--e1762},
  year={2023},
  publisher={Lippincott Williams \& Wilkins Hagerstown, MD}
}

@article{jiang2024large,
  title={Large brain model for learning generic representations with tremendous {EEG} data in BCI},
  author={Jiang, Wei-Bang and Zhao, Li-Ming and Lu, Bao-Liang},
  journal={arXiv preprint arXiv:2405.18765},
  year={2024}
}

@article{wang2024cbramod,
  title={Cbramod: A criss-cross brain foundation model for {EEG} decoding},
  author={Wang, Jiquan and Zhao, Sha and Luo, Zhiling and Zhou, Yangxuan and Jiang, Haiteng and Li, Shijian and Li, Tao and Pan, Gang},
  journal={arXiv preprint arXiv:2412.07236},
  year={2024}
}

@article{zhou2025csbrain,
  title={CSBrain: A Cross-scale Spatiotemporal Brain Foundation Model for {EEG} Decoding},
  author={Zhou, Yuchen and Wu, Jiamin and Ren, Zichen and Yao, Zhouheng and Lu, Weiheng and Peng, Kunyu and Zheng, Qihao and Song, Chunfeng and Ouyang, Wanli and Gou, Chao},
  journal={arXiv preprint arXiv:2506.23075},
  year={2025}
}

@article{chen2024toward,
  title={Toward reliable signals decoding for electroencephalogram: A benchmark study to {EEGNeX}},
  author={Chen, Xia and Teng, Xiangbin and Chen, Han and Pan, Yafeng and Geyer, Philipp},
  journal={Biomedical Signal Processing and Control},
  volume={87},
  pages={105475},
  year={2024},
  publisher={Elsevier}
}

@article{hu2022lora,
  title={Lora: Low-rank adaptation of large language models.},
  author={Hu, Edward J and Shen, Yelong and Wallis, Phillip and Allen-Zhu, Zeyuan and Li, Yuanzhi and Wang, Shean and Wang, Lu and Chen, Weizhu and others},
  journal={ICLR},
  volume={1},
  number={2},
  pages={3},
  year={2022}
}

@article{kemp2000analysis,
  title={Analysis of a sleep-dependent neuronal feedback loop: the slow-wave microcontinuity of the {EEG}},
  author={Kemp, Bob and Zwinderman, Aeilko H and Tuk, Bert and Kamphuisen, Hilbert AC and Oberye, Josefien JL},
  journal={IEEE Transactions on Biomedical Engineering},
  volume={47},
  number={9},
  pages={1185--1194},
  year={2000},
  publisher={IEEE}
}

@article{schalk2004bci2000,
  title={BCI2000: a general-purpose brain-computer interface (BCI) system},
  author={Schalk, Gerwin and McFarland, Dennis J and Hinterberger, Thilo and Birbaumer, Niels and Wolpaw, Jonathan R},
  journal={IEEE Transactions on biomedical engineering},
  volume={51},
  number={6},
  pages={1034--1043},
  year={2004},
  publisher={IEEE}
}

@article{goldberger2000physiobank,
  title={PhysioBank, PhysioToolkit, and PhysioNet: components of a new research resource for complex physiologic signals},
  author={Goldberger, Ary L and Amaral, Luis AN and Glass, Leon and Hausdorff, Jeffrey M and Ivanov, Plamen Ch and Mark, Roger G and Mietus, Joseph E and Moody, George B and Peng, Chung-Kang and Stanley, H Eugene},
  journal={circulation},
  volume={101},
  number={23},
  pages={e215--e220},
  year={2000},
  publisher={Lippincott Williams \& Wilkins}
}

@article{brunner2008bci,
  title={BCI Competition 2008--Graz data set A},
  author={Brunner, Clemens and Leeb, Robert and M{\"u}ller-Putz, Gernot and Schl{\"o}gl, Alois and Pfurtscheller, Gert},
  journal={Institute for knowledge discovery (laboratory of brain-computer interfaces), Graz University of Technology},
  volume={16},
  number={1-6},
  pages={34},
  year={2008}
}

@misc{inria-bci-challenge,
    author = {Jérémie Mattout and Manu and maucle and Wendy Kan},
    title = {BCI Challenge @ NER 2015},
    year = {2014},
    howpublished = {\url{https://kaggle.com/competitions/inria-bci-challenge}},
    note = {Kaggle}
}

@article{mumtaz2016mdd,
  title={MDD patients and healthy controls {EEG} data (new)},
  author={Mumtaz, Wajid},
  journal={figshare, Dataset},
  year={2016}
}

@inproceedings{harati2015improved,
  title={Improved {EEG} event classification using differential energy},
  author={Harati, Amir and Golmohammadi, Meysam and Lopez, Silvia and Obeid, Iyad and Picone, Joseph},
  booktitle={2015 IEEE Signal Processing in Medicine and Biology Symposium (SPMB)},
  pages={1--4},
  year={2015},
  organization={IEEE}
}

@article{rundo2019polysomnography,
  title={Polysomnography},
  author={Rundo, Jessica Vensel and Downey III, Ralph},
  journal={Handbook of clinical neurology},
  volume={160},
  pages={381--392},
  year={2019},
  publisher={Elsevier}
}

@article{maganti2013eeg,
  title={{EEG} and epilepsy monitoring},
  author={Maganti, Rama K and Rutecki, Paul},
  journal={CONTINUUM: Lifelong Learning in Neurology},
  volume={19},
  number={3},
  pages={598--622},
  year={2013},
  publisher={LWW}
}

@article{daly2008brain,
  title={Brain--computer interfaces in neurological rehabilitation},
  author={Daly, Janis J and Wolpaw, Jonathan R},
  journal={The Lancet Neurology},
  volume={7},
  number={11},
  pages={1032--1043},
  year={2008},
  publisher={Elsevier}
}

@article{lotte2018review,
  title={A review of classification algorithms for {EEG}-based brain--computer interfaces: a 10 year update},
  author={Lotte, Fabien and Bougrain, Laurent and Cichocki, Andrzej and Clerc, Maureen and Congedo, Marco and Rakotomamonjy, Alain and Yger, Florian},
  journal={Journal of neural engineering},
  volume={15},
  number={3},
  pages={031005},
  year={2018},
  publisher={iOP Publishing}
}

@article{simeoni2025dinov3,
  title={Dinov3},
  author={Sim{\'e}oni, Oriane and Vo, Huy V and Seitzer, Maximilian and Baldassarre, Federico and Oquab, Maxime and Jose, Cijo and Khalidov, Vasil and Szafraniec, Marc and Yi, Seungeun and Ramamonjisoa, Micha{\"e}l and others},
  journal={arXiv preprint arXiv:2508.10104},
  year={2025}
}

@article{brown2020language,
  title={Language models are few-shot learners},
  author={Brown, Tom and Mann, Benjamin and Ryder, Nick and Subbiah, Melanie and Kaplan, Jared D and Dhariwal, Prafulla and Neelakantan, Arvind and Shyam, Pranav and Sastry, Girish and Askell, Amanda and others},
  journal={Advances in neural information processing systems},
  volume={33},
  pages={1877--1901},
  year={2020}
}

@inproceedings{radford2023robust,
  title={Robust speech recognition via large-scale weak supervision},
  author={Radford, Alec and Kim, Jong Wook and Xu, Tao and Brockman, Greg and McLeavey, Christine and Sutskever, Ilya},
  booktitle={International conference on machine learning},
  pages={28492--28518},
  year={2023},
  organization={PMLR}
}

@article{xu2025lsm,
  title={LSM-2: Learning from Incomplete Wearable Sensor Data},
  author={Xu, Maxwell A and Narayanswamy, Girish and Ayush, Kumar and Spathis, Dimitris and Liao, Shun and Tailor, Shyam A and Metwally, Ahmed and Heydari, A Ali and Zhang, Yuwei and Garrison, Jake and others},
  journal={arXiv preprint arXiv:2506.05321},
  year={2025}
}

@article{narayanswamy2024scaling,
  title={Scaling wearable foundation models},
  author={Narayanswamy, Girish and Liu, Xin and Ayush, Kumar and Yang, Yuzhe and Xu, Xuhai and Liao, Shun and Garrison, Jake and Tailor, Shyam and Sunshine, Jake and Liu, Yun and others},
  journal={arXiv preprint arXiv:2410.13638},
  year={2024}
}

@article{baevski2020wav2vec,
  title={wav2vec 2.0: A framework for self-supervised learning of speech representations},
  author={Baevski, Alexei and Zhou, Yuhao and Mohamed, Abdelrahman and Auli, Michael},
  journal={Advances in neural information processing systems},
  volume={33},
  pages={12449--12460},
  year={2020}
}

@article{wei2021finetuned,
  title={Finetuned language models are zero-shot learners},
  author={Wei, Jason and Bosma, Maarten and Zhao, Vincent Y and Guu, Kelvin and Yu, Adams Wei and Lester, Brian and Du, Nan and Dai, Andrew M and Le, Quoc V},
  journal={arXiv preprint arXiv:2109.01652},
  year={2021}
}

@inproceedings{radford2021learning,
  title={Learning transferable visual models from natural language supervision},
  author={Radford, Alec and Kim, Jong Wook and Hallacy, Chris and Ramesh, Aditya and Goh, Gabriel and Agarwal, Sandhini and Sastry, Girish and Askell, Amanda and Mishkin, Pamela and Clark, Jack and others},
  booktitle={International conference on machine learning},
  pages={8748--8763},
  year={2021},
  organization={PmLR}
}

@article{sugden2023remote,
  title={Remote collection of electrophysiological data with brain wearables: opportunities and challenges},
  author={Sugden, Richard James and Pham-Kim-Nghiem-Phu, Viet-Linh Luke and Campbell, Ingrid and Leon, Alberto and Diamandis, Phedias},
  journal={Bioelectronic Medicine},
  volume={9},
  number={1},
  pages={12},
  year={2023},
  publisher={Springer}
}

@inproceedings{kommineni2024knowledge,
  title={Knowledge-guided eeg representation learning},
  author={Kommineni, Aditya and Avramidis, Kleanthis and Leahy, Richard and Narayanan, Shrikanth},
  booktitle={2024 46th Annual International Conference of the IEEE Engineering in Medicine and Biology Society (EMBC)},
  pages={1--6},
  year={2024},
  organization={IEEE}
}

@article{avramidis2025neural,
  title={Neural Codecs as Biosignal Tokenizers},
  author={Avramidis, Kleanthis and Feng, Tiantian and Jeong, Woojae and Lee, Jihwan and Cui, Wenhui and Leahy, Richard M and Narayanan, Shrikanth},
  journal={arXiv preprint arXiv:2510.09095},
  year={2025}
}

@article{looney2012ear,
  title={The in-the-ear recording concept: User-centered and wearable brain monitoring},
  author={Looney, David and Kidmose, Preben and Park, Cheolsoo and Ungstrup, Michael and Rank, Mike Lind and Rosenkranz, Karin and Mandic, Danilo P},
  journal={IEEE pulse},
  volume={3},
  number={6},
  pages={32--42},
  year={2012},
  publisher={IEEE}
}

@article{abbaspourazad2023large,
  title={Large-scale training of foundation models for wearable biosignals},
  author={Abbaspourazad, Salar and Elachqar, Oussama and Miller, Andrew C and Emrani, Saba and Nallasamy, Udhyakumar and Shapiro, Ian},
  journal={arXiv preprint arXiv:2312.05409},
  year={2023}
}

@inproceedings{zhang2024low,
  title={Low-resource vision challenges for foundation models},
  author={Zhang, Yunhua and Doughty, Hazel and Snoek, Cees GM},
  booktitle={Proceedings of the IEEE/CVF Conference on Computer Vision and Pattern Recognition},
  pages={21956--21966},
  year={2024}
}

@article{dong2023efficient,
  title={Efficient adaptation of large vision transformer via adapter re-composing},
  author={Dong, Wei and Yan, Dawei and Lin, Zhijun and Wang, Peng},
  journal={Advances in Neural Information Processing Systems},
  volume={36},
  pages={52548--52567},
  year={2023}
}

@inproceedings{elizalde2023clap,
  title={Clap learning audio concepts from natural language supervision},
  author={Elizalde, Benjamin and Deshmukh, Soham and Al Ismail, Mahmoud and Wang, Huaming},
  booktitle={ICASSP 2023-2023 IEEE International Conference on Acoustics, Speech and Signal Processing (ICASSP)},
  pages={1--5},
  year={2023},
  organization={IEEE}
}

@article{bendr,
  title={BENDR: Using transformers and a contrastive self-supervised learning task to learn from massive amounts of EEG data},
  author={Kostas, Demetres and Aroca-Ouellette, Stephane and Rudzicz, Frank},
  journal={Frontiers in Human Neuroscience},
  volume={15},
  pages={653659},
  year={2021},
  publisher={Frontiers Media SA}
}

@inproceedings{neurogpt,
  title={Neuro-gpt: Towards a foundation model for eeg},
  author={Cui, Wenhui and Jeong, Woojae and Th{\"o}lke, Philipp and Medani, Takfarinas and Jerbi, Karim and Joshi, Anand A and Leahy, Richard M},
  booktitle={2024 IEEE International Symposium on Biomedical Imaging (ISBI)},
  pages={1--5},
  year={2024},
  organization={IEEE}
}

@article{eegpt,
  title={Eegpt: Pretrained transformer for universal and reliable representation of eeg signals},
  author={Wang, Guangyu and Liu, Wenchao and He, Yuhong and Xu, Cong and Ma, Lin and Li, Haifeng},
  journal={Advances in Neural Information Processing Systems},
  volume={37},
  pages={39249--39280},
  year={2024}
}

@article{margaux2012objective,
  title={Objective and Subjective Evaluation of Online Error Correction during P300-Based Spelling},
  author={Margaux, Perrin and Emmanuel, Maby and S{\'e}bastien, Daligault and Olivier, Bertrand and J{\'e}r{\'e}mie, Mattout},
  journal={Advances in Human-Computer Interaction},
  volume={2012},
  number={1},
  pages={578295},
  year={2012},
  publisher={Wiley Online Library}
}

@article{obeid2016temple,
  title={The temple university hospital EEG data corpus},
  author={Obeid, Iyad and Picone, Joseph},
  journal={Frontiers in neuroscience},
  volume={10},
  pages={196},
  year={2016},
  publisher={Frontiers Media SA}
}

@article{lee2025large,
  title={Are Large Brainwave Foundation Models Capable Yet? Insights from Fine-tuning},
  author={Lee, Na and Barmpas, Konstantinos and Panagakis, Yannis and Adamos, Dimitrios and Laskaris, Nikolaos and Zafeiriou, Stefanos},
  journal={arXiv preprint arXiv:2507.01196},
  year={2025}
}

@article{yang2023biot,
  title={Biot: Biosignal transformer for cross-data learning in the wild},
  author={Yang, Chaoqi and Westover, M and Sun, Jimeng},
  journal={Advances in Neural Information Processing Systems},
  volume={36},
  pages={78240--78260},
  year={2023}
}

@article{GramfortEtAl2013a,
  title = {{{MEG}} and {{EEG}} Data Analysis with {{MNE}}-{{Python}}},
  author = {Gramfort, Alexandre and Luessi, Martin and Larson, Eric and Engemann, Denis A. and Strohmeier, Daniel and Brodbeck, Christian and Goj, Roman and Jas, Mainak and Brooks, Teon and Parkkonen, Lauri and H{\"a}m{\"a}l{\"a}inen, Matti S.},
  year = {2013},
  volume = {7},
  pages = {1--13},
  doi = {10.3389/fnins.2013.00267},
  journal = {Frontiers in Neuroscience},
  number = {267}
}

@article{paszke2019pytorch,
  title={Pytorch: An imperative style, high-performance deep learning library},
  author={Paszke, Adam and Gross, Sam and Massa, Francisco and Lerer, Adam and Bradbury, James and Chanan, Gregory and Killeen, Trevor and Lin, Zeming and Gimelshein, Natalia and Antiga, Luca and others},
  journal={Advances in neural information processing systems},
  volume={32},
  year={2019}
}

@article{wang2025eegmamba,
  title={Eegmamba: An eeg foundation model with mamba},
  author={Wang, Jiquan and Zhao, Sha and Luo, Zhiling and Zhou, Yangxuan and Li, Shijian and Pan, Gang},
  journal={Neural Networks},
  pages={107816},
  year={2025},
  publisher={Elsevier}
}

@inproceedings{
jiang2025neurolm,
title={Neuro{LM}: A Universal Multi-task Foundation Model for Bridging the Gap between Language and {EEG} Signals},
author={Weibang Jiang and Yansen Wang and Bao-liang Lu and Dongsheng Li},
booktitle={The Thirteenth International Conference on Learning Representations},
year={2025},
url={https://openreview.net/forum?id=Io9yFt7XH7}
}

@article{ma2025codebrain,
  title={CodeBrain: Towards Decoupled Interpretability and Multi-Scale Architecture for EEG Foundation Model},
  author={Ma, Jingying and Wu, Feng and Lin, Qika and Xing, Yucheng and Liu, Chenyu and Jia, Ziyu and Feng, Mengling},
  journal={arXiv preprint arXiv:2506.09110},
  year={2025}
}

@article{liu2025echo,
  title={ECHO: Toward Contextual Seq2Seq Paradigms in Large EEG Models},
  author={Liu, Chenyu and Deng, Yuqiu and Liu, Tianyu and Zhou, Jinan and Zhou, Xinliang and Jia, Ziyu and Ding, Yi},
  journal={arXiv preprint arXiv:2509.22556},
  year={2025}
}

\appendix

\section{Referenced Datasets}\label{apd:datasets}
\subsection{Motor Imagery}
Motor Imagery classification tasks include identifying a user's intention to move specific parts of the body. Robust motor imagery classification could allow for development of neuro-assistive technologies.

\textbf{Physionet MI}~\citep{schalk2004bci2000, goldberger2000physiobank}: consists of 1500 1-2 minute EEG recordings from 109 participants, for four  real and imagined motor tasks (open and close left fist, right fist, both fists and both feet). EEG recordings consist of 64 channels and were sampled at 160Hz. Each trial was segmented into 4-second windows and all experiments were performed in a 5-fold between subjects cross validation setting, such that each subject was in the test fold once. Balanced Accuracy, Cohen's Kappa and F1-Macro are reported.

\textbf{BCI IV Competition 2A}~\citep{brunner2008bci}: Motor Imagery Classification dataset consists of data recorded from 9 participants across two sessions recorded on different days. Data were collected using 22 EEG channels at 250Hz for four imagined motor movements (left arm, right arm, both feet and tongue). Trials are segmented into 4 second chunks and Leave-One-Subject-Out cross validation is performed for all classification experiments using this dataset. This is done to evaluate models ability to generalize across subjects.
\subsection{Error Related Negativity}
\textbf{Kaggle ERN}~\citep{inria-bci-challenge}: The dataset includes EEG recordings from 26 participants who perform tasks using an online P300 speller interface, and is primarily used to study event-related potentials related to erroneous responses. The EEG data were collected using 56 EEG electrodes and were downsampled to 200 Hz. The classification task is to detect when the selected item is not the intended.

\subsection{Mental Health}
\textbf{MDD MAL}~\citep{mumtaz2016mdd}: Depression classification task from EEG recorded at resting state, composed of eyes closed and eyes open conditions. Each condition is recorded for 5 minutes with 19 electrodes at 200Hz sampling rate. The dataset consists of 34 patients diagnosed with major depressive disorder and 30 healthy controls. For classification, non-overlapping 10s windows are considered as input samples to the models. Evaluation is performed in a between subject 5-fold cross validation setup.

\subsection{Sleep Stages}
\textbf{Sleep EDFx}~\citep{kemp2000analysis}: contains 197 whole-night PolySomnoGraphic sleep recordings with EEG, EOG and chin EMG recorded at 100Hz and annotated for sleep stages every 30s. In this work, EEG electrodes Fpz-Cz and Pz-Oz are considered for analysis. The sleep stages are annotated for Wake, REM, Movement, NREM-1, NREM-2, NREM-3 and NREM-4. For comparability to prior studies, NREM-3 and NREM-4 have been combined to a single class yielding 5 classes. 5-fold between-subjects cross validation is performed for evaluation.

\subsection{Events}

\textbf{TUEV}~\citep{harati2015improved}: contains EEG dataset derived from Temple University EEG Corpus~\citep{obeid2016temple} with annotations for categorizing EEG segments into six classes: (1) spike and sharp wave (SPSW), (2) generalized periodic epileptiform discharges (GEPD), (3) periodic latealized epileptiform discharges (PLED), (4) eye movements (EYEM), (5) artifact and (6) background (BCKG). TUEV is recorded at 200Hz sampling rate and each annotation is segmented into 5s long EEG timeseries.

\section{Referenced Models}\label{apd:models}

\textbf{EEGNet}~\citep{lawhern2018eegnet} is a lightweight convolution-based model that combines temporal and spatial convolution layers followed by a classification head. To accommodate for the varying sampling rates of the datasets, the kernel size of the temporal convolution layer and separable convolution layer had been set to half the sampling rate and one-eigth the sampling rate of the respective downstream dataset. To test the effects of scaling model sizes within EEGNet, three versionswere defined: (1) \texttt{EEGNet}: F1=8, D=2; (2) \texttt{EEGNet large}: F1=16, D=4; (3) \texttt{EEGNet Huge}: F1=32, D=8, where F! is the number of temporal filters and D is the depth multiplier.

\textbf{EEGNeX}~\citep{chen2024toward} is an improved version of EEGNet architecture that employs strided convolutions to increase the effective temporal window that the model is able to capture, thereby providing performance improvements on BCI tasks.

\textbf{SparcNet}~\citep{jing2023development}: 1D CNN model that leverages dense residual connections to capture spatiotemporal relations in EEG signals effectively.

\textbf{LaBraM}~\citep{jiang2024large} is an EEG foundation model with a pre-training objective based on masked token prediction. The underlying neural tokenizer is trained with large-scale EEG data through patching the EEG time series into tokens.

\textbf{CBraMod}~\citep{wang2024cbramod} is a foundation model trained on 25,000 hours pre-training data that aims to model temporal and spatial characteristics through distinct (criss-cross) attention mechanisms. This model employs a patch-based masked reconstruction scheme for pre-training.

\textbf{CSBrain}~\citep{zhou2025csbrain} is an attention-based foundation model for EEG decoding with novel cross-scale spatiotemporal tokenization and structured sparse attention. It is pre-trained using masked reconstruction objectives.

\section{Implementation Details}
\label{apd:training}

All experiments were conducted in Python 3.13. Pre-processing utilized the MNE-Python~\citep{GramfortEtAl2013a}. PyTorch~\citep{paszke2019pytorch} was used to build and train all deep learning models. All experiments were conducted on a cluster of 4 RTX A6000 GPUs.

All PEFT experiments utilized LoRa with rank $r=4$. 
Gradient clipping with value 1.0 was used for full fine-tuning experiments.
AdamW optimizer with learning rates between [1e-2, 1e-3] were used for linear probe, [1e-3, 1e-4] for LoRA and [2e-4, 1e-5] for full-finetuning experiments.
Cosine Anneal Learning rate scheduler with 10\% epochs as warmup steps was used across all experiments. 
For Fully supervised models, a learning rate plateau scheduler was used.
All models were trained for up to 30 epochs for Physionet MI, BCIC IV-2A, Kaggle ERN and MDD MAL datasets, and for 50 epochs for Sleep EDF and TUEV datasets.
The model checkpoint with the lowest overall validation loss was used for evaluation of downstream performance. 
Further details on hyperparameters are provided with the accompanying ~\href{https://github.com/Uncertain-Quark/eegfm_eval_toolkit}{Github repository}.
\par 
Parameter counts for Linear probe and LoRA corresponding to each dataset and model are as reported in Table~\ref{tab:model_params}
\begin{table*}[]
\begin{tabular}{lcccccc}
\toprule
\textbf{Model} & \textbf{Kaggle ERN} & \textbf{Physionet MI} & \textbf{BCIC IV 2A} & \textbf{TUEV} & \textbf{MDD MAL} & \textbf{Sleep EDF} \\
\midrule
LaBraM(LP)     & 23k                 & 205k                  & 71k                 & 103k          & 76k              & 61k                \\
LaBraM(LoRA)   & 118k                & 301k                  & 167k                & 200k          & 172k             & 157k               \\
CBraMod(LP)    & 22k                 & 204k                  & 70k                 & 102k          & 76k              & 60k                \\
CBraMod(LoRA)  & 137k                & 320k                  & 185k                & 217k          & 191k             & 175k               \\
CSBrain(LP)    & 22k                 & 205k                  & 70k                 & 102k          & 76k              & 60k                \\
CSBrain(LoRA)  & 157k                & 340k                  & 205k                & 236k          & 210k             & 195k               \\
\bottomrule
\end{tabular}
\caption{Model parameters for LoRA and Linear probe (LP) settings across foundation models for evaluation datasets.}
\label{tab:model_params}
\end{table*}

\section{Metrics}
\label{apd:metrics}
Consistent with prior EEG self-supervised modeling works~\citep{zhou2025csbrain, kommineni2024knowledge, avramidis2025neural, wang2024cbramod, jiang2024large}, we use the following evaluation metrics:
\begin{itemize}
    \item \textbf{Balanced Accuracy (BAC)}: Computes the unweighted average recall across classes which is able to better account for class imbalances in datasets compared to raw accuracy scores.
    \begin{equation*}
        \text{BAC} = \frac{1}{K}\Sigma_{j=1}^{K}\frac{TP_j}{TP_j + FN_j}
    \end{equation*}
    where $K$ is the number of classes, $TP_j$ is the number of True Positives and $FN_j$ is the number of False Negatives for class $j$.
    \item \textbf{Cohen's Kappa ($\kappa$)}: Measures the mutual agreement between model predictions and true labels while accounting for agreement by chance. A value of 0 indicates no agreement between the model predictions and true labels, whereas 1.0 indicates perfect alignment.
    \begin{equation*}
        \kappa = \frac{p_o - p_e}{1 - p_{e}}
    \end{equation*}
    where $p_o$ is the observed agreement and $p_e$ is the hypothetical agreement by chance.
    \item \textbf{Area Under the Receiver Operating Curve (AUROC)}: Computes the area under the True Positive Rate (TPR) versus False Positive Rate (FPR) curve, computed at different thresholds for classification. An AUROC of 0.5 thus indicates random-chance performance, whereas a value of 1.0 indicates perfect classification.
    \item \textbf{F1-Macro}: Computes unweighted average of F1 scores across all classes in dataset. Since the average is unweighted, this metric is more robust to class imbalances.
    \begin{equation*}
        \text{F1-Macro} = \frac{1}{K}\Sigma_{j=1}^{K}\frac{2.\text{Pr}_j.\text{Re}_j}{\text{Pr}_j + \text{Re}_j}
    \end{equation*}
    where $K$ is the number of classes, $\text{Pr}_j$ and $\text{Re}_j$ refer to precision and recall of class $j$.
\end{itemize}

\section{CbraMod \& LaBraM Sample Efficiency}
\begin{figure*}
    \centering
    \includegraphics[width=\linewidth]{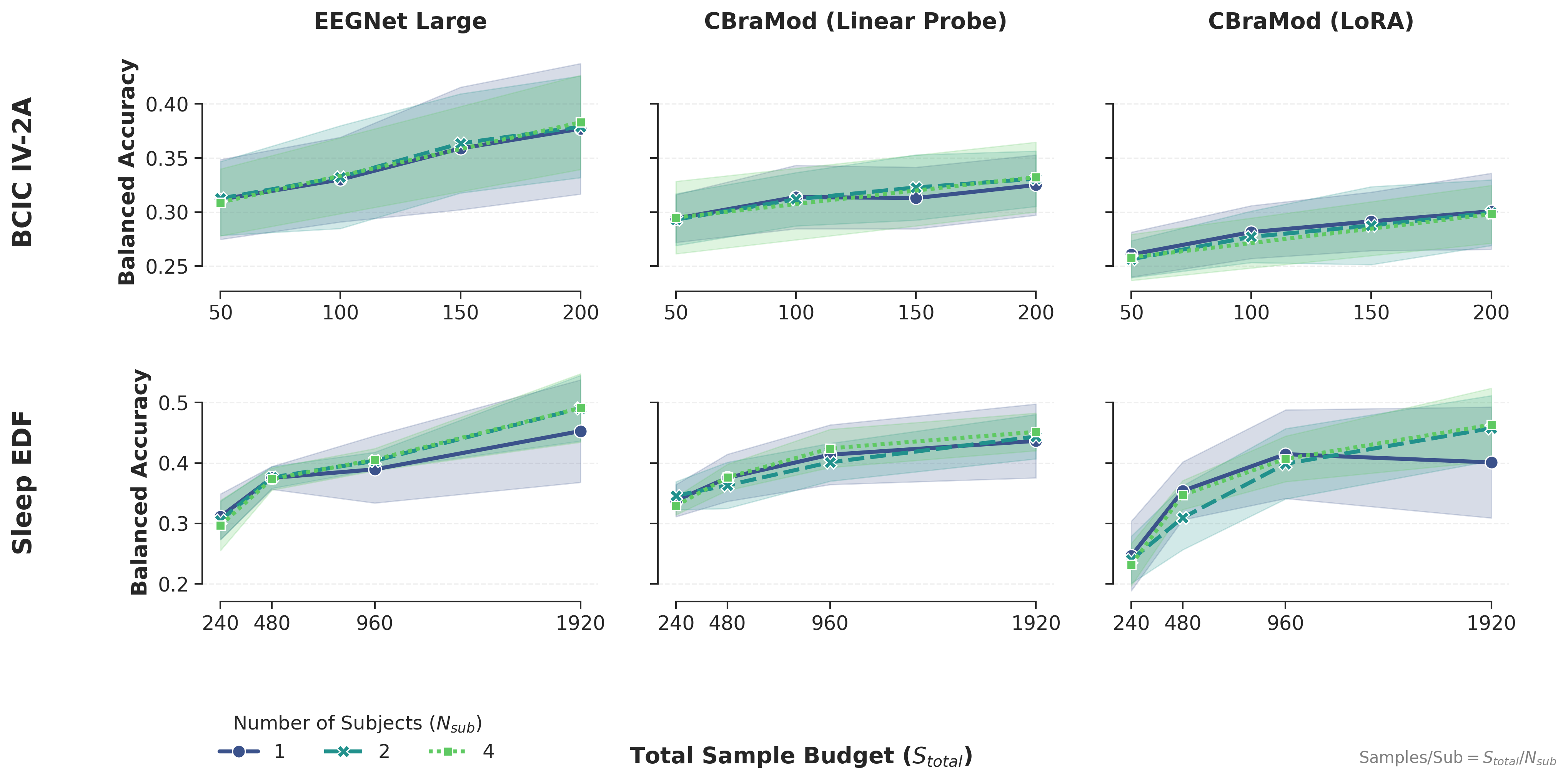}
    \caption{Total Budget sampling plot for CBramod under linear probe and LoRA settings compared to EEGNet Large}
    \label{fig:sample_efficiency_cbramod}
\end{figure*}

\begin{figure*}
    \centering
    \includegraphics[width=\linewidth]{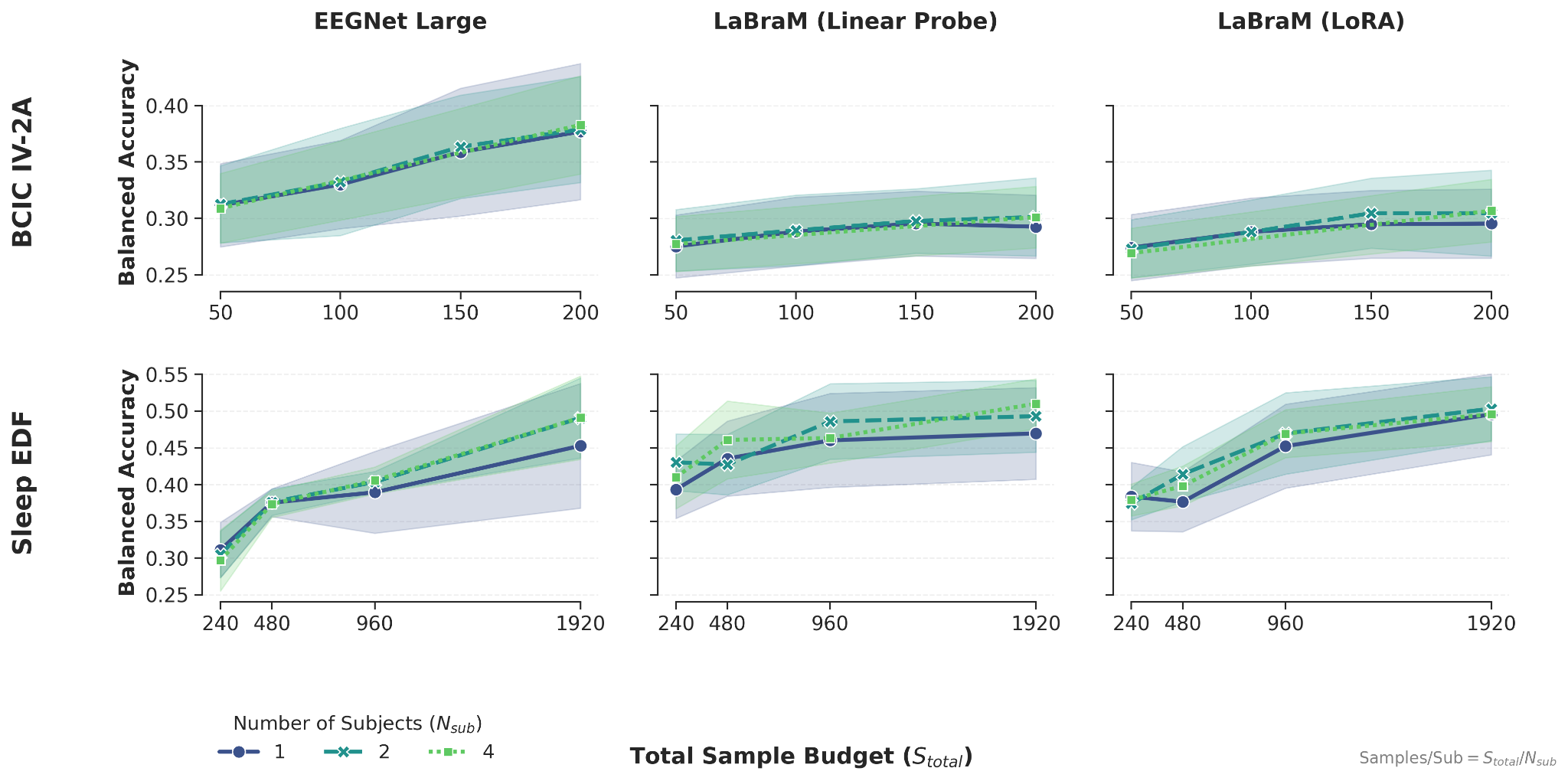}
    \caption{Total Budget sampling plot for LaBraM under linear probe and LoRA settings compared to EEGNet Large.}
    \label{fig:sample_efficiency_labram}
\end{figure*}

Sampling Efficiency plot for CBraMod (Fig~\ref{fig:sample_efficiency_cbramod}) and LaBraM (Fig~\ref{fig:sample_efficiency_labram}) in comparison with EEGNet.

\section{Full Data Results}
\label{apd:full_results}
Classification results for each evaluation dataset on all the listed models can be found below:
\begin{itemize}
    \item Physionet MI: Table~\ref{tab:physionet_mi_results}
    \item BCIC IV-2A: Table~\ref{tab:bciciv_2a_results}
    \item Kaggle ERN: Table~\ref{tab:kaggle_ern_results}
    \item MDD MAL: Table~\ref{tab:mdd_mal_results}
    \item Sleep EDF: Table~\ref{tab:sleep_edf_results}
    \item TUEV: Table~\ref{tab:tuev_results}
\end{itemize}

\begin{table*}[]
\centering
\begin{tabular}{l c c c}
\toprule
\textbf{Model} & \textbf{BAC} & \textbf{F1-Macro} & ${\boldsymbol{\kappa}}$ \\
\midrule
\texttt{Supervised Models} \\
\quad EEGNet & $61.34 \pm 1.91$ & $61.40 \pm 1.96$ & $48.45 \pm 2.54$ \\
\quad EEGNet Large & $61.85 \pm 1.95$ & $61.83 \pm 2.02$ & $49.14 \pm 2.59$ \\
\quad EEGNet Huge & $61.74 \pm 1.75$ & $61.70 \pm 1.81$ & $49.00 \pm 2.34$ \\
\quad EEGNeX & {\boldsymbol{$65.58 \pm 1.73$}} & {\boldsymbol{$65.65 \pm 1.82$}} & {\boldsymbol{$54.13 \pm 2.30$}} \\
\quad SparcNet & $62.02 \pm 1.54$ & $61.85 \pm 1.57$ & $49.37 \pm 2.05$ \\
\midrule
\texttt{LaBraM} \\
\quad Full-finetune & $57.25 \pm 1.55$ & $57.22 \pm 1.62$ & $43.02 \pm 2.05$ \\
\quad Linear Probe & $48.82 \pm 1.31$ & $48.74 \pm 1.30$ & $31.77 \pm 1.77$ \\
\quad LoRA & $49.10 \pm 1.32$ & $49.06 \pm 1.30$ & $32.16 \pm 1.74$ \\
\midrule
\texttt{CBraMod} \\
\quad Full-finetune & $64.02 \pm 2.52$ & $63.96 \pm 2.47$ & $52.06 \pm 3.36$ \\
\quad Linear Probe & $51.97 \pm 1.63$ & $51.84 \pm 1.53$ & $35.98 \pm 2.17$ \\
\quad LoRA & $62.05 \pm 2.52$ & $61.94 \pm 2.51$ & $49.41 \pm 3.36$ \\
\midrule
\texttt{CSBrain} \\ 
\quad Full-finetune & $62.34 \pm 2.33$ & $62.22 \pm 2.38$ & $49.80 \pm 3.10$ \\
\quad Linear Probe & $51.81 \pm 1.66$ & $51.59 \pm 1.52$ & $35.77 \pm 2.21$ \\
\quad LoRA & $58.95 \pm 1.78$ & $58.90 \pm 1.72$ & $45.28 \pm 2.37$ \\
\bottomrule
\end{tabular}
\caption{Classification results for Physionet MI dataset for Supervised and Foundation Models. Balanced Accuracy (BAC), F1-Macro and Cohens Kappa ($\kappa$) metrics are reported. Best value under each metric is reported in bold.}
\label{tab:physionet_mi_results}
\end{table*}

\begin{table*}[]
\centering
\begin{tabular}{l c c c}
\toprule
\textbf{Model} & \textbf{BAC} & \textbf{F1-Macro} & ${\boldsymbol{\kappa}}$ \\
\midrule
\texttt{Supervised Models} \\
\quad EEGNet & $54.96 \pm 7.05$ & $53.64 \pm 7.32$ & $39.94 \pm 9.40$ \\
\quad EEGNet Large & $58.14 \pm 7.07$ & $56.84 \pm 7.78$ & $44.19 \pm 9.43$ \\
\quad EEGNet Huge & $58.76 \pm 6.60$ & $\boldsymbol{57.86 \pm 7.20}$ & $45.01 \pm 8.80$ \\
\quad EEGNeX &  $\boldsymbol{59.10 \pm 11.25}$ & $56.93 \pm 13.38$ & $\boldsymbol{45.47 \pm 15.00}$ \\
\quad SparcNet & $56.96 \pm 10.43$ & $55.09 \pm 12.31$ & $42.62 \pm 13.91$ \\
\midrule
\texttt{LaBraM} \\
\quad Full-finetune & $50.58 \pm 9.20$ & $49.01 \pm 10.22$ & $34.10 \pm 12.26$ \\
\quad Linear Probe & $45.25 \pm 5.47$ & $43.62 \pm 6.15$ & $27.01 \pm 7.29$ \\
\quad LoRA & $42.57 \pm 5.47$ & $41.09 \pm 6.06$ & $23.43 \pm 7.29$ \\
\midrule
\texttt{CBraMod} \\
\quad Full-finetune & $56.15 \pm 8.77$ & $54.91 \pm 9.51$ & $41.54 \pm 11.69$ \\
\quad Linear Probe &$43.81 \pm 5.18$ & $42.89 \pm 5.35$ & $25.08 \pm 6.91$ \\
\quad LoRA &  $51.37 \pm 11.95$ & $48.08 \pm 14.32$ & $35.16 \pm 15.93$ \\
\midrule
\texttt{CSBrain} \\ 
\quad Full-finetune & $55.27 \pm 8.39$ & $54.18 \pm 8.82$ & $40.35 \pm 11.19$ \\
\quad Linear Probe & $46.51 \pm 6.82$ & $44.88 \pm 8.19$ & $28.68 \pm 9.09$ \\
\quad LoRA & $55.83 \pm 6.46$ & $54.61 \pm 6.91$ & $41.10 \pm 8.61$ \\
\bottomrule
\end{tabular}
\caption{Classification results for BCI Competition IV-2A dataset for Supervised and Foundation Models. Balanced Accuracy (BAC), F1-Macro and Cohens Kappa ($\kappa$) metrics are reported. Best value under each metric is reported in bold.}
\label{tab:bciciv_2a_results}
\end{table*}

\begin{table*}[]
\centering
\begin{tabular}{l c c c}
\toprule
\textbf{Model} & \textbf{BAC} & \textbf{AUROC} & ${\boldsymbol{\kappa}}$ \\
\midrule
\texttt{Supervised Models} \\
\quad EEGNet & $62.74 \pm 2.09$ & $65.06 \pm 1.83$ & $27.76 \pm 4.39$ \\
\quad EEGNet Large & $64.94 \pm 2.23$ & $68.41 \pm 1.80$ & $32.43 \pm 4.70$ \\
\quad EEGNet Huge& $64.54 \pm 1.48$ & $68.12 \pm 1.02$ & $31.29 \pm 3.06$ \\
\quad EEGNeX  & $\boldsymbol{65.00 \pm 1.38}$ & $\boldsymbol{70.69 \pm 1.19}$ & $\boldsymbol{33.49 \pm 2.55}$ \\
\quad SparcNet  & $61.70 \pm 1.30$ & $66.49 \pm 2.30$ & $24.47 \pm 3.14$ \\
\midrule
\texttt{LaBraM} \\
\quad Full-finetune & $58.51 \pm 1.79$ & $64.39 \pm 1.92$ & $18.66 \pm 2.81$ \\
\quad Linear Probe & $56.11 \pm 0.97$ & $60.89 \pm 2.01$ & $13.15 \pm 2.15$ \\
\quad LoRA& $58.25 \pm 1.04$ & $63.06 \pm 1.59$ & $17.20 \pm 2.58$ \\
\midrule
\texttt{CBraMod} \\
\quad Full-finetune & $61.47 \pm 1.25$ & $68.21 \pm 1.43$ & $24.49 \pm 2.44$ \\
\quad Linear Probe& $57.72 \pm 0.57$ & $64.34 \pm 1.06$ & $17.33 \pm 1.32$ \\
\quad LoRA & $60.71 \pm 1.02$ & $66.36 \pm 1.20$ & $22.24 \pm 1.91$ \\
\midrule
\texttt{CSBrain} \\ 
\quad Full-finetune& $63.19 \pm 1.60$ & $70.42 \pm 1.99$ & $28.16 \pm 2.30$ \\
\quad Linear Probe & $60.97 \pm 1.19$ & $65.48 \pm 0.81$ & $21.76 \pm 2.84$ \\
\quad LoRA & $61.55 \pm 1.20$ & $67.49 \pm 1.53$ & $24.53 \pm 1.89$ \\
\bottomrule
\end{tabular}
\caption{Classification results for Kaggle ERN dataset for Supervised and Foundation Models. Balanced Accuracy (BAC), AUROC and Cohens Kappa ($\kappa$) metrics are reported. Best value under each metric is reported in bold.}
\label{tab:kaggle_ern_results}
\end{table*}

\begin{table*}[]
\centering
\begin{tabular}{l c c c}
\toprule
\textbf{Model} & \textbf{BAC} & \textbf{AUROC} & ${\boldsymbol{\kappa}}$ \\
\midrule
\texttt{Supervised Models} \\
\quad EEGNet & $86.89 \pm 3.88$ & $93.17 \pm 3.33$ & $72.39 \pm 7.97$ \\
\quad EEGNet Large  & $84.98 \pm 6.82$ & $90.98 \pm 6.65$ & $68.83 \pm 13.14$ \\
\quad EEGNet Huge & $83.61 \pm 7.21$ & $90.64 \pm 4.28$ & $66.03 \pm 13.78$ \\
\quad EEGNeX  & $86.21 \pm 8.36$ & $93.36 \pm 5.08$ & $71.44 \pm 16.67$ \\
\quad SparcNet  & $79.32 \pm 6.22$ & $91.10 \pm 5.92$ & $57.74 \pm 12.03$ \\
\midrule
\texttt{LaBraM} \\
\quad Full-finetune  & $88.72 \pm 3.12$ & $95.61 \pm 2.75$ & $76.61 \pm 6.71$ \\
\quad Linear Probe & $87.24 \pm 4.68$ & $93.78 \pm 4.35$ & $73.81 \pm 9.08$ \\
\quad LoRA& $87.88 \pm 4.30$ & $93.67 \pm 4.15$ & $75.12 \pm 9.16$ \\
\midrule
\texttt{CBraMod} \\
\quad Full-finetune & $83.08 \pm 6.69$ & $91.72 \pm 6.25$ & $64.94 \pm 12.66$ \\
\quad Linear Probe& $80.71 \pm 5.43$ & $87.73 \pm 7.42$ & $60.72 \pm 11.29$ \\
\quad LoRA & $83.95 \pm 10.10$ & $92.16 \pm 7.74$ & $67.59 \pm 19.89$ \\
\midrule
\texttt{CSBrain} \\ 
\quad Full-finetune& $88.81 \pm 5.41$ & $95.19 \pm 5.30$ & $76.69 \pm 10.76$ \\
\quad Linear Probe & $88.03 \pm 5.46$ & $95.40 \pm 4.00$ & $75.48 \pm 11.11$ \\
\quad LoRA & $\boldsymbol{89.34 \pm 5.54}$ & $\boldsymbol{96.08 \pm 4.03}$ & $\boldsymbol{77.99 \pm 11.05}$ \\
\bottomrule
\end{tabular}
\caption{Classification results for supervised and foundation models trained on depression classification task using MDD MAL dataset. Balanced Accuracy (BAC), AUROC and Cohens Kappa ($\kappa$) metrics are reported. Best value under each metric is reported in bold.}
\label{tab:mdd_mal_results}
\end{table*}

\begin{table*}[]
\centering
\begin{tabular}{l c c c}
\toprule
\textbf{Model} & \textbf{BAC} & \textbf{F1-Macro} & ${\boldsymbol{\kappa}}$ \\
\midrule
\texttt{Supervised Models} \\
\quad EEGNet & $70.20 \pm 1.29$ & $64.58 \pm 2.22$ & $71.94 \pm 2.43$ \\
\quad EEGNet Large & $71.61 \pm 1.20$ & $65.87 \pm 2.26$ & $73.51 \pm 1.71$ \\
\quad EEGNet Huge & $71.14 \pm 1.36$ & $66.53 \pm 1.02$ & $74.14 \pm 1.64$ \\
\quad EEGNeX  & $70.31 \pm 1.22$ & $69.36 \pm 1.03$ & $77.55 \pm 2.21$ \\
\quad SparcNet  & $71.01 \pm 2.64$ & $68.52 \pm 2.73$ & $75.41 \pm 2.87$ \\
\midrule
\texttt{LaBraM} \\
\quad Full-finetune & $72.86 \pm 1.22$ & $69.88 \pm 2.24$ & $75.09 \pm 3.22$ \\
\quad Linear Probe & $64.85 \pm 0.50$ & $63.91 \pm 0.83$ & $69.88 \pm 1.43$ \\
\quad LoRA & $65.96 \pm 1.86$ & $64.56 \pm 2.36$ & $71.16 \pm 2.16$ \\
\midrule
\texttt{CBraMod} \\
\quad Full-finetune & $\boldsymbol{74.58 \pm 0.85}$ & $\boldsymbol{73.34 \pm 0.39}$ & $\boldsymbol{79.63 \pm 1.36}$ \\
\quad Linear Probe & $57.71 \pm 1.19$ & $59.50 \pm 1.28$ & $69.48 \pm 0.80$ \\
\quad LoRA & $69.61 \pm 2.22$ & $67.06 \pm 1.72$ & $75.42 \pm 0.86$ \\
\midrule
\texttt{CSBrain} \\ 
\quad Full-finetune & $71.17 \pm 1.17$ & $70.16 \pm 0.63$ & $76.25 \pm 2.30$ \\
\quad Linear Probe &  $63.44 \pm 2.17$ & $63.11 \pm 1.20$ & $67.39 \pm 1.91$ \\
\quad LoRA & $72.73 \pm 0.98$ & $71.38 \pm 1.69$ & $77.80 \pm 2.05$ \\
\bottomrule
\end{tabular}
\caption{Classification results for Sleep EDF dataset for Supervised and Foundation Models. Balanced Accuracy (BAC), F1-Macro and Cohens Kappa ($\kappa$) metrics are reported. Best value under each metric is reported in bold.}
\label{tab:sleep_edf_results}
\end{table*}

\begin{table*}[]
\centering
\begin{tabular}{l c c c}
\toprule
\textbf{Model} & \textbf{BAC} & \textbf{F1-Macro} & ${\boldsymbol{\kappa}}$ \\
\midrule
\texttt{Supervised Models} \\
\quad EEGNet & $51.67 \pm 2.10$ & $53.62 \pm 1.22$ & $\boldsymbol{54.29\pm4.42}$ \\
\quad EEGNet Large & $53.27 \pm 3.88$ & $\boldsymbol{54.33 \pm 3.01}$ & $50.58 \pm 3.19$ \\
\quad EEGNet Huge &  $52.58\pm1.82$ & $51.62\pm2.00$ & $49.31\pm2.01$ \\
\quad EEGNeX  & $47.50\pm3.01$ & $48.55\pm2.87$ & $52.07\pm2.93$ \\
\quad SparcNet  & $52.10 \pm 3.51$ & $49.66 \pm 2.15$ & $50.63 \pm 4.06$ \\
\midrule
\texttt{LaBraM} \\
\quad Full-finetune & $\boldsymbol{57.58 \pm 1.92}$ & $53.01 \pm 3.43$ & $50.83 \pm 4.89$ \\
\quad Linear Probe & $42.81 \pm 2.35$ & $38.39 \pm 1.64$ & $34.71 \pm 2.42$ \\
\quad LoRA & $43.94\pm2.53$ & $39.50\pm 1.73$ & $35.17 \pm 3.63$ \\
\midrule
\texttt{CBraMod} \\
\quad Full-finetune & $54.70 \pm 2.34$ & $50.46 \pm 1.74$ & $50.85 \pm 1.65$ \\
\quad Linear Probe &$37.40 \pm 1.60$ & $36.66 \pm 2.58$ & $36.91 \pm 3.53$ \\
\quad LoRA & $52.77 \pm 1.57$ & $50.55 \pm 1.32$ & $47.49 \pm 2.38$ \\
\midrule
\texttt{CSBrain} \\ 
\quad Full-finetune & $51.88 \pm 2.16$ & $45.38 \pm 3.55$ & $44.38 \pm 4.16$ \\
\quad Linear Probe &  $47.35 \pm 3.15$ & $42.54 \pm 2.79$ & $41.76 \pm 4.94$ \\
\quad LoRA & $52.41 \pm 2.61$ & $47.71 \pm 2.82$ & $50.47 \pm 2.10$ \\
\bottomrule
\end{tabular}
\caption{Classification results for TUEV dataset for Supervised and Foundation Models. Balanced Accuracy (BAC), F1-Macro and Cohens Kappa ($\kappa$) metrics are reported. Best value under each metric is reported in bold.}
\label{tab:tuev_results}
\end{table*}

\section{Sampling Rate Ablations}
\label{apd:sampling_rate_ablations_appendix}
Sampling rate ablations for supervised model performance between native dataset and the resampled data to match preprocessing steps of foundation models. Results are as reported in Table~\ref{tab:sampling_rate_ablations}.

\begin{table*}[]
\begin{tabular}{lccccc}
\toprule
\textbf{Model}    & \textbf{Sampling Rate} & \textbf{Physionet MI} & \textbf{BCIC IV-2A} & \textbf{MDD MAL} & \textbf{Sleep-EDF} \\
\midrule
\multirow{2}{*}{EEGNet Large} & 200 Hz        & 62.8                 & 57.3               & 84.5             & 69.7              \\
                              & Native        & 61.9                 & 58.1               & 85.0             & 71.6              \\
\multirow{2}{*}{EEGNet Small} & 200 Hz        & 62.5                 & 54.2               & 83.1             & 68.8              \\
                              & Native        & 61.3                 & 54.9               & 86.8             & 70.2              \\
\multirow{2}{*}{EEGNet Huge}  & 200 Hz        & 61.5                 & 60.2               & 84.5             & 68.9              \\
                              & Native        & 61.7                 & 58.7               & 83.6             & 71.1\\  
\bottomrule
\end{tabular}
\caption{Sampling rate ablations for model performance between native sampling rate of datasets compared to sampling rate of foundation models (200Hz). Results indicate no noticeable impact of sampling rate on the model performance.}
\label{tab:sampling_rate_ablations}
\end{table*}

\section{MDD MAL Sample Efficiency Results}
\label{apd:mdd_mal_sample_efficiency}
Sample efficiency results for MDD MAL dataset are reported in Table~\ref{tab:performance_comparison}
\begin{table*}[htbp]
\centering
\begin{tabular}{llcccc}
\toprule
& & \textbf{Supervised} & \multicolumn{3}{c}{\textbf{Foundation Models}} \\
\cmidrule(lr){3-3} \cmidrule(lr){4-6}
\textbf{Setting} & \textbf{Budget} & \textbf{EEGNet} & \textbf{LaBraM} & \textbf{CBraMod} & \textbf{CSBrain} \\
\midrule
\multirow{5}{*}{Linear Probe (LP)} 
& 40  & 49.1 (8.8)  & 63.1 (16.5) & 60.3 (11.8) & 69.1 (12.7) \\
& 80  & 56.2 (10.1) & 72.7 (13.7) & 68.3 (8.5)  & 67.1 (19.3) \\
& 160 & 66.2 (12.6) & 80.6 (11.6) & 73.1 (10.5) & 77.0 (11.9) \\
& 320 & 78.1 (4.9)  & 86.7 (5.1)  & 75.6 (7.4)  & 82.8 (9.7)  \\
& 640 & 84.6 (6.0)  & 88.0 (4.8)  & 77.9 (5.7)  & 86.4 (6.3)  \\
\midrule
\multirow{5}{*}{LoRA} 
& 40  & 49.1 (8.8)  & 62.7 (13.7) & 60.4 (13.6) & 61.1 (17.2) \\
& 80  & 56.2 (10.1) & 69.6 (18.0) & 62.5 (13.1) & 75.1 (10.0) \\
& 160 & 66.2 (12.6) & 79.5 (11.1) & 67.4 (11.7) & 76.5 (12.5) \\
& 320 & 78.1 (4.9)  & 83.7 (9.0)  & 76.9 (14.0) & 82.5 (8.4)  \\
& 640 & 84.6 (6.0)  & 88.6 (4.3)  & 81.7 (9.6)  & 88.4 (5.9)  \\
\bottomrule
\end{tabular}
\caption{Sample efficiency results of Linear Probe (LP) and LoRA setting for MDD MAL dataset. Results
indicate that EEG foundation models provide noticeable performance gains compared to supervised models under low samples for long window tasks. EEGNet does not have Linear Probe or LoRA it is fully supervised and the results are copied to make comparison easier for linear probe and LoRA experiments.}
\label{tab:performance_comparison}
\end{table*}
\section{MDD MAL Channel Efficiency}
Additional channel efficiency results for MDD MAL dataset are reported in Table~\ref{tab:channel_efficiency_mdd_mal}.
\begin{table*}[htbp]
\centering
\begin{tabular}{@{}llccccc@{}}
\toprule
\textbf{Model}           & \textbf{PEFT} & \textbf{Central} & \textbf{Frontal} & \textbf{Midline} & \textbf{1 channel per lobe} & \textbf{2 channels per lobe} \\ \midrule
EEGNet Large             & -             & 84.5 (7.3)       & 83.8 (2.9)       & 84.5 (6.8)       & 85.8 (7.0)                  & 84.1 (6.5)                   \\ \addlinespace
\multirow{2}{*}{LaBraM}  & LP            & 83.5 (3.2)       & 86.0 (6.7)       & 85.7 (2.7)       & 89.7 (2.5)                  & 88.7 (3.2)                   \\
                         & LoRA          & 84.2 (2.0)       & 86.0 (8.5)       & 85.2 (2.1)       & 89.0 (3.4)                  & 89.4 (2.9)                   \\ \addlinespace
\multirow{2}{*}{CBraMod} & LP            & 72.2 (10.8)      & 76.4 (6.4)       & 73.8 (12.2)      & 73.6 (11.2)                 & 75.0 (7.7)                   \\
                         & LoRA          & 84.2 (7.4)       & 85.8 (9.3)       & 87.1 (3.6)       & 85.0 (7.1)                  & 83.3 (9.5)                   \\ \addlinespace
\multirow{2}{*}{CSBrain} & LP            & 77.2 (9.2)       & 82.2 (8.9)       & 79.2 (9.0)       & 78.3 (9.8)                  & 85.4 (4.7)                   \\
                         & LoRA          & 81.8 (4.1)       & 82.5 (9.9)       & 84.1 (7.6)       & 85.5 (7.5)                  & 87.5 (4.8)                   \\ \bottomrule
\end{tabular}
\caption{Channel efficiency experiments for MDD MAL dataset for supervised model EEGNet and foundation models (LaBraM, CBraMod and CSBrain) under linear probe and LoRA settings.}
\label{tab:channel_efficiency_mdd_mal}
\end{table*}

\end{document}